\documentclass{article}

\usepackage{arxiv}
\usepackage{amsmath,amssymb,amsthm}
\usepackage[utf8]{inputenc} 
\usepackage[T1]{fontenc}    
\usepackage{url}            
\usepackage{booktabs}       
\usepackage{amsfonts}       
\usepackage{nicefrac}       
\usepackage{microtype}      
\usepackage{atbegshi}%
\usepackage[utf8]{inputenc}
\usepackage{subcaption}
\usepackage{csquotes}
\usepackage{xcolor}
\usepackage{graphicx}
\usepackage{booktabs}
\usepackage{algorithm}
\usepackage[noend]{algorithmic}
\usepackage{float}
\usepackage{soul}
\usepackage{natbib}

\usepackage{tikz}
\usepackage{pgfplots}
\pgfplotsset{compat=1.17}

\sethlcolor{lightgray}

\DeclareMathOperator*{\argmax}{arg\,max}

\newcommand{\COMM}[1]{\hfill\textcolor{gray!80}{// #1}}

\title{Computing Diverse Sets of High Quality TSP Tours by EAX-Based Evolutionary Diversity Optimisation}

\author{
 Adel Nikfarjam \\
Optimisation and Logistics\\School of Computer Science\\The University of Adelaide\\
  \texttt{adel.nikfarjam@adelaide.edu.au} \\
   \And
 Jakob Bossek \\
Statistics and Optimization\\Dept. of Information Systems\\University of M\"unster\\
  \texttt{jakob.bossek@wi.uni-muenster.de} \\
  \And
 Aneta Neumann \\
Optimisation and Logistics\\School of Computer Science\\The University of Adelaide\\
  \texttt{aneta.neumann@adelaide.edu.au} \\
    \And
 Frank Neumann \\
Optimisation and Logistics\\School of Computer Science\\The University of Adelaide\\
  \texttt{frank.neumann@adelaide.edu.au} \\}
\begin{document}
\maketitle
\begin{abstract}
Evolutionary algorithms based on edge assembly crossover~(EAX) constitute some of the best performing incomplete solvers for the well-known traveling salesperson problem~(TSP). Often, it is desirable to compute not just a single solution for a given problem, but a diverse set of high quality solutions from which a decision maker can choose one for implementation.
Currently, there are only a few approaches for computing a diverse solution set for the TSP. Furthermore, almost all of them assume that the optimal solution is known. 
In this paper, we introduce evolutionary diversity optimisation~(EDO) approaches for the TSP that find a diverse set of tours when the optimal tour is known or unknown. We show how to adopt EAX to not only find a high-quality solution but also to maximise the diversity of the population. The resulting EAX-based EDO approach, termed EAX-EDO is capable of obtaining diverse high-quality tours when the optimal solution for the TSP is known or unknown. A comparison to existing approaches shows that they are clearly outperformed by EAX-EDO.
\end{abstract}


\section{Introduction}
\label{sec:introduction}

Classically, the aim of optimisation is to find an optimal solution or a high-quality solution for a given problem. In recent years, the evolutionary computation community has shown increasing interest in finding a diverse set of solutions for optimisation problems~\cite{ArulkumaranCT19,FontaineLSSTH19,MouretM20}.
Being provided with a diverse set of good solutions enables decision-makers to choose between different alternatives. In the context of the traveling salesperson problem~(TSP) for instance, decision-makers may choose to visit a city earlier or avoid an edge if they have different options with reasonable price. Moreover, if predefined assumptions or input parameters change slightly and the chosen solution becomes infeasible, decision-makers are able to switch to other still feasible solutions in the set. Last but not least, maximising diversity of tours subject to a quality threshold provides us with invaluable information about the solution space; given that it is difficult to comprehend solution spaces in combinatorial optimisation. For instance, we can realize which edges are more/less costly to be replaced in optimal tours. Several applications of using a diverse set of solution can be found in the literature, e.g. in personalized agent planning~\cite{roberts2012using} and air traffic control~\cite{bloem2014air}.

\subsection{Related Work}

In the literature, several studies can be found aiming to compute a diverse set of solutions based on constrained programming and automated planning. \citet{hebrard2005finding} introduced a framework to compute diverse sets of solutions for constrained programming; it was adopted in many studies~\cite{srivastava2007domain,coman2011generating,vadlamudi2016combinatorial,katz2020reshaping}.
In most approaches, linear-time greedy algorithms are used to explore the solution space. This limits not only the completeness of algorithms but also their efficiency within a time-bound \cite{vadlamudi2016combinatorial}. In contrast to the aforementioned papers, this study aims to use evolutionary algorithms~(EAs) to obtain a diverse set of tours for the well-known NP-hard traveling salesperson problem~(TSP). Several algorithms have been proposed to solve the TSP \cite{helsgaun2000effective,lin1973effective,nagata2013powerful,xie2008multiagent} where EAX by \citet{nagata2013powerful} is widely known as one of the state-of-the-art inexact solvers.

Evolutionary diversity optimisation~(EDO) was first introduced in~\cite{ulrich2011maximizing}. Here, the objective is to find a set of solutions which all have an acceptable quality but differ in one or more predefined features. In recent years, many of the studies in the context of EDO focused on generating diverse sets of images and benchmark instances for the TSP \cite{alexander2017evolution,doi:10.1162/evcoa00274}. They generated sets of similar images differing in aesthetics and TSP instances with regard to the characteristics of the problem that make an instance easy or difficult to solve for different TSP-solvers. 

In terms of different diversity measures, studies evolving diverse sets of TSP instances and sets of images by using the star-discrepancy measure~\cite{neumann2018discrepancy} and indicators from multi-objective optimisation \cite{neumann2019evolutionary} have been carried out recently. Specific mutation operators to achieve high diversity in TSP instances without using diversity preservation mechanisms explicitly have been introduced in \cite{bossek2019evolving}.
There are a few papers utilising the context of EDO to generate diverse sets of solutions for classical combinatorial optimization problems. \citet{bossek2021evolutionary} investigated the Minimum Spanning Tree problem in the context of EDO. They proved that for a small set including two solutions diversity can be obtained in polynomial time. \citet{neumann2021diversifying} computed diverse sets of solutions for submodular optimisation problems with uniform and knapsack constraints. They first introduced a sampling-based greedy algorithm to obtain diversity; then, they present an EDO approach to improve the results. To the best of our knowledge only \citet{viet2020evolving, 10.1145/3449639.3459313, nikfarjam2021entropy} investigated EDO to compute diverse sets of tours for the TSP. \citet{viet2020evolving} introduced two distance-based diversity measures and studied simple EAs with $k$-OPT neighborhood search operators exclusively. This is while \citet{nikfarjam2021entropy} employed an entropy-based diversity measure focusing on diversifying segments of solutions. They both worked under the assumption of an optimal tour being known a-priori.

\subsection{Our Contribution}
In this study, we adopt EAX~\cite{nagata2013powerful} in the context of evolutionary diversity optimisation and introduce an approach called EAX-EDO in order to obtain high-quality tours for the TSP while maximising diversity of the population simultaneously.
In classical EAs typically a loss of diversity in the population can be observed when improving the quality of solutions.
Although EAX benefits from an entropy-based diversity preservation mechanism, it merely focuses on avoiding premature convergence rather than maximising the diversity of the final population. 
We adopt the state-of-the-art EAX crossover operator for solving TSP instances and introduce a modification called EAX-EDO crossover which is tailored towards simultaneous optimisation of solution quality and population diversity. We incorporate the EAX-EDO crossover into three EDO approaches. In the case of unknown optimal solutions, we introduce two algorithms. First, we adopt a two-stage framework from the literature of EDO which alternates between phases of optimising the cost and optimising the diversity (the two-stage EAX-EDO). Second, we introduce the single-stage EAX-EDO designed to simultaneously optimise both quality of solutions and population diversity. Moreover, we conduct a comparison between the frameworks with the classical EAX and the exact Gurobi optimiser.
Our experimental investigations show that EAX-EDO is capable of generating solutions with very decent quality. More importantly, EAX-EDO can maintain and even increase the diversity of the population during the process of optimising the quality of solutions. This is while EAX requires to sacrifice the diversity of population to gain solutions of higher quality. Moreover, we conduct a series of experiments to examine the robustness of the four competitors' populations against minor changes. The outcome indicates the single-EAX-EDO's superiority in terms of populations' robustness.   

The remainder of the paper is structured as follows. In Section~\ref{sec:problem_description}, we formally introduce the considered diversity setting and introduce a diversity measure based on entropy. Afterwards, we present the EAX-EDO crossover in Section~\ref{sec:EAX_EDO} and evolutionary algorithms for the case where an optimal solution is known (Section~\ref{sec:EAX_EDO_known}) or unknown~(Section~\ref{sec:EAX_EDO_unknown}). In Section~\ref{sec:experiments}, we conduct a series of experiments showing the advantage of EAX-EDO when the optimal solution for the TSP is known and unknown on TSP benchmark instances. Finally, we finish with some concluding remarks.


\section{Problem description}
\label{sec:problem_description}

The TSP problem is defined on a directed complete graph $G=(V, E)$ where $V$ is a set of nodes of size $n=|V|$ and $E$ is a set of pairwise edges between the nodes. 
There is a non-negative weight (distance) $w(e)$ associated with each edge $e = (u,v) \in E$. Throughout the paper, we assume that TSP instances are symmetric, i.e. $w((u,v))=w((v,u))$ holds for all $u,v \in V$.
The goal is to find a permutation $p : V \to V$ that minimises the cost function
$$c(p) = w(p(n),p(1)) + \sum_{i=1}^{n-1} w(p(i),p(i+1)).$$
Our evolutionary algorithms maintain (multi-)sets $P = \{p_1, \ldots, p_{\mu}\}$ of $\mu$ permutations. In the following, it is often helpful to identify a permutation/tour $p = (p(1), \ldots, p(n))$ by means of its edge set $E(p)=\{(p(1), p(2)), (p(2), p(1)) \ldots, (p(n),p(1)), (p(1), p(n))\}.$

We study the TSP in the context of evolutionary diversity optimisation. Let $G$ be a TSP instance and let $OPT$ be the value of an optimal salesperson tour of $G$. Given a parameter $\alpha > 0$, the goal is to evolve a population $P$ of $\mu$ tours such that (1) all tours adhere to a minimal solution quality threshold, i.e. $c(p) \leq c_{\max}$ for all $p \in P$, where $c_{\max} = (1+\alpha)OPT$ and simultaneously (2) some measure $D(P)$ is maximized that quantifies the diversity of the tours in the population. 
We denote the diversity of $P$ by $D(P)$ and state the diversity optimisation problem as

\begin{align*}
    & \max D(P) &\\
   & \text{subject to} &\\
    & c(p) \leq (1+\alpha)OPT & \forall p \in P.
\end{align*}


\subsection{Diversity Measures}
\citet{viet2020evolving} were the first to study the TSP in the context of EDO. For this purpose, the authors introduced two distance-based diversity measures which they termed \emph{edge diversity}~($ED$) and \emph{pairwise diversity}~($PD$).

$ED$ is the mean pairwise distance between individuals of the population
\begin{equation*}
ED(P) = \sum_{p \in P}\sum_{q \in P} |E(p)\setminus E(q)|.
\end{equation*}
In consequence its value is maximised by equalizing the occurrences of edges within the population. 

On the other hand, $PD$ focuses on uniform pairwise distance between the individuals and is defined as
\begin{equation*}
    PD(P) = \frac{1}{n\mu} \sum_{p \in P} \min_{q \in P\setminus{\{p\}}} \{|E(p)\setminus E(q)|\}.
\end{equation*}
In our study, we adopt a diversity measure based on the well-known information-theoretic concept of entropy since the entropy measure is shown to yield superior diversity in benchmarks compared distance-based measures \cite{nagata2020high}. The overall entropy of a given population $P$ is defined as
\begin{align*}
H(P) = \sum_{e \in E} h(e) \text{ with } h(e)=-\left(\frac{f(e)}{2n\mu}\right)\ln{\left(\frac{f(e)}{2n\mu}\right)}.
\end{align*}
where $h(e)$ is the contribution of an edge $e \in E$ to the entropy.
Here, $f(e)$ is the number of individuals 
in $P$ that use edge $e \in E$. 
Note that $2n\mu$ is the total number of edges in the population as each of the $\mu$ permutations consists of exactly $2n$ edges. The contribution of the edges with $f(e)=0$ to the population is equal to zero. The entropy of a population $P$ is minimal if and only if $P$ includes $\mu$ copies of a single tour. In this setting the minimum entropy value is
\begin{align*}
    H_{min} = -2n\left(\frac{1}{2n}\ln{\left(\frac{1}{2n}\right)}\right) = \ln(2n).\label{eq:mininal_entropy}
\end{align*}
\citet{nikfarjam2021entropy} showed that the maximum entropy value occurs when all edges appear roughly the same number of times in the population.


\section{EAX-EDO Crossover}
\label{sec:EAX_EDO}

EAX crossover (EAX CO) is a permutation-based crossover operator which is used in a genetic algorithm~(GA) named EAX as well. It was first introduced by~\citet{nagata1997edge}. Several versions of EAX can be found in the literature \cite{nagata2013powerful,nagata2020high,nagata2006new}. The version introduced by \citet{nagata2013powerful} constitutes a key component in one of the best performing incomplete algorithms for solving the TSP. 
It is shown that the algorithm is cable of obtaining the optimal or best-known solutions for most benchmark instances and more importantly improved 11 best-known solutions.
The GA works in two stages. The first stage which is the main part of the algorithm uses EAX-1AB for crossover, while EAX-Block2 crossover is utilised in the second stage. While EAX-1AB results in offspring very similar to one of the parents, EAX-Block2 brings about more different offspring. Having the first stage converged and incapable of improving the objective function any further, the second stage is initialised to explore solution areas for possible improvements. The algorithm also benefits from an entropy-based diversity mechanism. 
\begin{algorithm}[t!
]
\begin{algorithmic}[1]
\REQUIRE{Two parent tours $p_1$ and $p_2$.}
\STATE Derive an AB-cycle from $p_1$ and $p_2$
\STATE Construct an intermediate tour from $p_1$ by removing the edges of $E(p_1)$ and adding edges of $E(p_2)$ in the AB-cycle.
\STATE Count the number of sub-tours in the intermediate solution and store it in $n_{sub}$.
\WHILE{$n_{sub} > 2$} 
\STATE Connect two sub-tours based on the neighbourhood search A and update $n_{sub}$. 
\ENDWHILE
\IF{$n_{sub} = 2$}
\STATE Connect two sub-tours based on the neighbourhood search B.
\ENDIF
\end{algorithmic}
\caption{EAX-EDO Crossover} 
\label{alg:EAX-EDO}
\end{algorithm}

During each iteration, the GA generates $n_{ch}$ tours from the same parents. Among those tours, it chooses the tours improving the cost of the solution compared to its parents as well as the entropy score of the population simultaneously if there is such a tour; otherwise, it selects the tour with minimum loss of entropy per improvement of the cost. It is crucial to notice that entropy is used to merely avoid premature convergence and not for the sake of a structurally diverse final population. We refer the interested to \cite{nagata2013powerful} for more details about the variants of EAX and the diversity preservation mechanism.      

\begin{figure*}[ht]
    \centering
    \includegraphics[width=\textwidth]{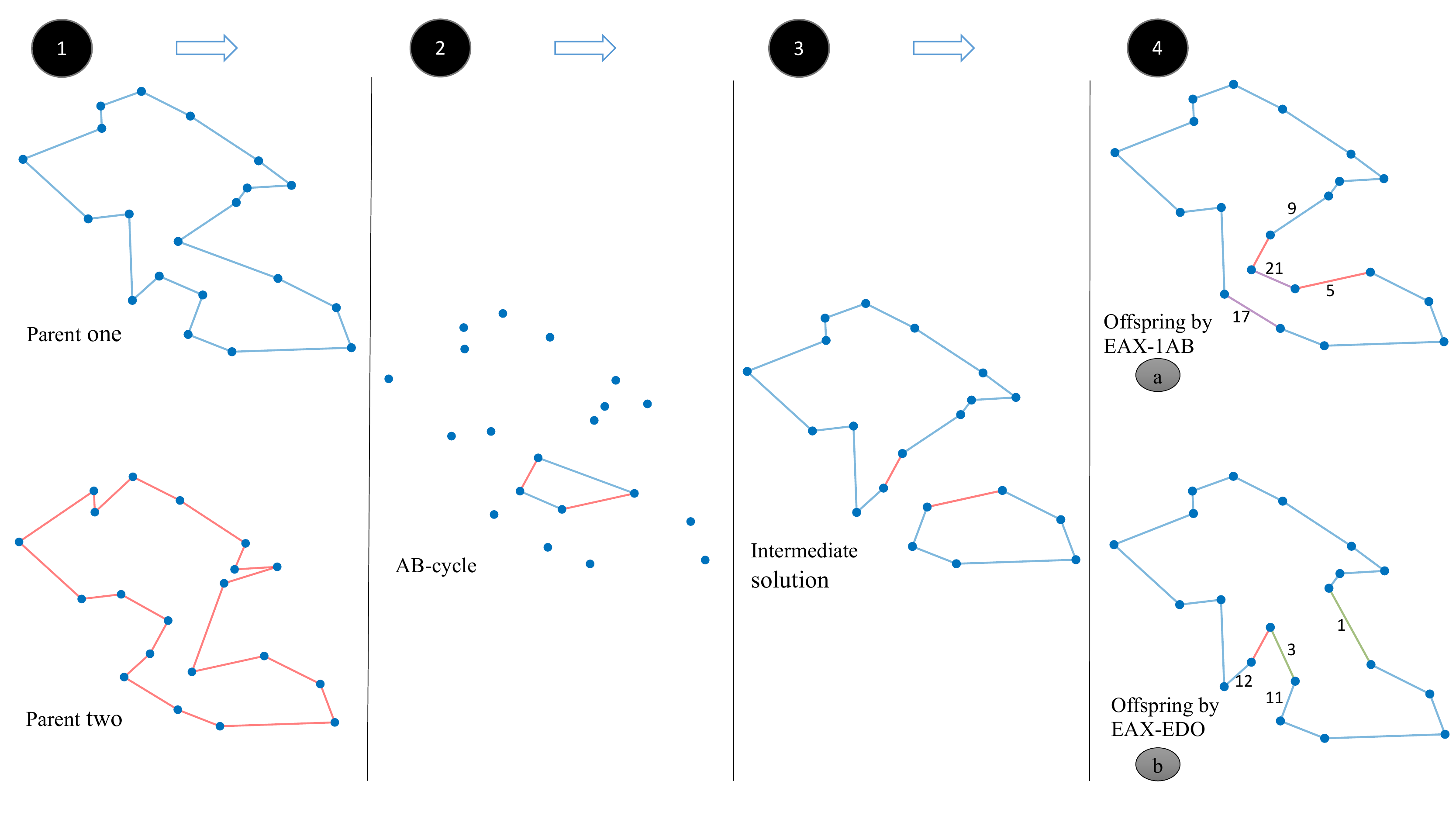}
    \caption{The representation of the steps to implement EAX-1AB and EAX-EDO. Up to step three the process of implementation for both crossover operators is the same. In step four, EAX-1AB (top) constructs the shortest tour possible, while EAX-EDO (bottom) generates a tour contributing to entropy the most while employing quality threshold (the numbers correspond to occurrences of associated edges in an imaginary population).}
    \label{fig:example_eax_1ab_vs_eax_edo}
\end{figure*}

In this study, we utilise EAX-1AB due to efficiency and simplicity of this version compared to other variants. EAX-1AB consists of three steps.
Let $p_1$ and $p_2$ be two parents selected from the population, and $E(p_1)$ and $E(p_2)$ be the sets of edges forming $p_1$ and $p_2$, respectively.
Firstly, an AB-cycle is derived from $p_1$ and $p_2$. An AB-cycle (see Figure~\ref{fig:example_eax_1ab_vs_eax_edo}.2) is a cycle where edges of $E(p_1)$ and $E(p_2)$ are linked, alternately. To form an AB-cycle, we start from a random node $v$. Then, we randomly select an edge $(v, u)$ from $E(p_1)$ and set $v=u$. In a similar manner, we add another edge to the tracing path from $p_2$ going through $v$, and reset $v$. We continue tracing nodes between $p_1$ and $p_2$ until an AB-cycle is formed in the trace path. Next, an intermediate solution $t$ (see Figure~\ref{fig:example_eax_1ab_vs_eax_edo}.3) is constructed from $p_1$ by adding edges of $E(p_2)$ and removing edges of $E(p_1)$ in the AB-cycle. Finally, a tour is generated by connecting all sub-tours of the intermediate solution (see Figure~\ref{fig:example_eax_1ab_vs_eax_edo}.4). Note that some AB-cycles are formed by two overlapping edges, one from $P_1$, the other from $p_2$. Such an AB-cycle is ineffective because it results in an intermediate solution, same as $p_1$. In this case, discard the ineffective AB-cycle from the tracing path, set $v$ to the last node in the tracing path, and this time select the other edge going through $v$ (there are always two edges going through $v$).

Algorithm \ref{alg:EAX-EDO} outlines EAX-EDO Crossover (EAX-EDO CO). The difference between EAX-EDO CO and EAX-1AB is the last step where the sub-tours are connected into a valid TSP tour (compare Figure~\ref{fig:example_eax_1ab_vs_eax_edo}.4.a (The Above) and Figure~\ref{fig:example_eax_1ab_vs_eax_edo}.4.b). In EAX, the sub-tour $r$ with the minimum number of edges is selected, and connected to another sub-tour $r'$ by removing an edge from each of them and adding two new edges. For this purpose, $4$-tuples of edges are selected such that $\{e_1, e_2, e_3, e_4\} = \arg \max \{-w(e_1)-w(e_2)+w(e_3)+w(e_4)\}$ where $e_1 \in E(r)$ and  $e_2 \in E(t) \setminus E(r)$ . Where $E(r)$ and $E(t)$ denotes the set of edges formed sub-tour $r$ and the intermediate solution $t$. For the sake of reduced computational cost, the search is limited in the way that either end of $e_3$ should be among the $N_{near}$ nearest nodes to either end of $e_3$ (here, $N_{near}$ is set to 10). we refer these steps as neighborhood search~A. In EAX-EDO CO, the neighbourhood search~A is implemented until two sub-tours are left. Then, neighbourhood search~B is started. First, all possible 4-tuples of edges complying $c(r)+c(r')-w(e_1)-w(e_2)+w(e_3)+w(e_4) \leq c_{max}$ are stored. From all the possible candidates, the 4-tuple of edges is selected where $\{e_1, e_2, e_3, e_4\} = \arg \max \{\Delta h(e_1)+\Delta h(e_2)+\Delta h(e_3)+\Delta h(e_4)\}.$
Here, $\Delta h(e_i)$ is the difference to the contribution of edge $i$ when it is either added or removed from the intermediate solution.


\section{EAX-EDO for Known Optimal Solution}
\label{sec:EAX_EDO_known}

A wide range of highly successful algorithms have been proposed in the literature to solve the TSP and optimal solutions can be obtained for a wide range of even large instances. In the following, we introduce an EAX-EDO approach that starts with an optimal solution and computes a diverse set of high quality solutions based on it. The algorithm is outlined in Algorithm~\ref{alg:diversity_maximizing_EA}.
The algorithm is initialised with $\mu$ copies of an optimal tour. Then, the entropy value of the population is calculated and stored in $H$. In this stage, $H$ should be equal to $H_{min}$. Having selected the parents, one offspring is generated by EAX-EDO~CO. If the cost of the offspring is at most $c_{max} = (1+\alpha)OPT$, it is added to $P$; then, an individual with $\arg \max H(P\setminus \{p\})$ is removed. Otherwise, the offspring is discarded. We repeat these steps until a termination criterion is met. 
Note that since the algorithm is initialized with $\mu$ copies of a single tour, a mutation operator should be used in initial iterations ($1\,000$ fitness evaluation in our experiments). This is because, EAX-EDO as a crossover operator requires two different parent solutions to generate a new offspring. For this purpose, we used 2-OPT in this study. 2-OPT is a random neighbourhood search. In 2-OPT, offspring is first formed by coping all the parent's edges. Then, having removed two edges randomly, we connect each end of the edges to the other edge in a way that it forms a complete TSP tour.     

\begin{algorithm}[t!]
\begin{algorithmic}[1]
\REQUIRE{Population $P$, minimal quality threshold $c_{max}$}
\WHILE{termination criterion is not met}
\STATE Choose $p_1$ and $p_2 \in P$ based on parent selection procedure, produce one offspring $p_3$ by crossover.\\ 
\IF{$c(p_3) \leq c_{max}$}
\STATE Add $p_3$ to $P$.
\ENDIF
\IF{$|P| = \mu+1$}
\STATE Remove one individual $p$ from $P$, where $p = \argmax_{q\in P} H(P \setminus \{q\})$.
\ENDIF
\ENDWHILE
\end{algorithmic}
\caption{Diversity-Maximising-EA}
\label{alg:diversity_maximizing_EA}
\end{algorithm}


\section{EAX-EDO for Unknown Optimal Solution}
\label{sec:EAX_EDO_unknown}

We now discuss the more general case where an optimal solution is unknown. In this case, it is important to diversify the set of solutions and increase the quality of the solution set. \citet{ulrich2011maximizing} introduced an approach to achieve both goals. In this approach, the algorithm switches between the cost minimisation and diversity maximisation. The approach is shown in Algorithm~\ref{alg:two_stage}. It starts with a cost minimising phase (see Algorithm~\ref{alg:cost_minimizing_phase}). The cost minimisation algorithm is initialised with a population optimised by 2-OPT in terms of cost (as it is done by the EAX genetic algorithm in \cite{nagata2013powerful}). Next, two individuals are selected to serve as the parents and one tour is generated by EAX-1AB~CO. The offspring replaces the first parent if it has a lower cost. Otherwise, it is discarded. These steps continue until an inner termination criterion is met for the cost optimisation stage. The worst found solution within the population dictates the least quality threshold $c_{max}$ for the diversity optimisation phase (see Algorithm~\ref{alg:diversity_maximizing_EA}). The algorithm switches between these stages until an overall termination criterion is met. Note that we switch cost minimisation phase off after $M$ consecutive failures in finding a shorter tour. There are, although, two disadvantages regarding this approach. First, several parameters need to be tuned to allocate the budget between the two phases. More importantly, diversity maximisation is neglected during cost minimising phase and vice versa. This causes a negative impact on the algorithm's efficiency.     
\begin{algorithm}[t!]
\begin{algorithmic}[1]
\REQUIRE{Initial Population $P$, and a limit for consecutive failures in improvement of the shortest tour $M$.}
\STATE Let $Best$ be the shortest tour in $P$.
\STATE $q \gets 0$ \COMM{$q$ number of consecutive failures in improvement of the shortest tour}
\WHILE{termination criterion is not met}
\IF{$q < M$}
\STATE $P \gets \text{Cost-Minimising-EA}(P)$. \COMM{Alg.~\ref{alg:cost_minimizing_phase}}
\STATE Let $c_{max}$ be the largest tour length in $P$.
\ENDIF
\IF{$c(p) < c(Best)$}
\STATE Update $Best$
\STATE $q \gets 0$
\ELSE
\STATE Set $q \gets q+1$
\ENDIF
\STATE $P \gets \text{Diversity-Maximising-EA}(P, c_{max})$ \COMM{Alg.~\ref{alg:diversity_maximizing_EA}}
\ENDWHILE
\end{algorithmic}
\caption{Two-stage EAX-EDO}
\label{alg:two_stage}
\end{algorithm}
\begin{algorithm}[t]
\begin{algorithmic}[1]
\REQUIRE{Population $P$}
\WHILE{termination criterion is not met}
\STATE Choose randomly two individuals, $p_1, p_2 \in P$, as the parents and generate on offspring $p_3$ by EAX-1AB\\
\IF{$c(p_3) \leq c(p_1)$}
\STATE Replace $p_1$ with $p_3$ in $P$.
\ENDIF
\ENDWHILE
\end{algorithmic}
\caption{Cost-Minimising-EA}
\label{alg:cost_minimizing_phase}
\end{algorithm}

Next, we introduce a single-stage algorithm to overcome the aforementioned issues. In this algorithm, two tours $p_1$ and $p_2$ are simultaneously generated such that $c(p_1) \leq c(p_2)$ and $H(P \setminus \{p_1\}) \leq H(P \setminus \{p_2\})$. I.e. $p_1$ is dedicated to cost optimisation, while $p_2$ is generated to increase the entropy of the population. Algorithm~\ref{alg:single_stage} outlines the approach by means of pseudo-code. The algorithm is initialised with a population of tours optimised locally by 2-OPT. Let $Best$ be the best found solution in $P$. Within the evolutionary loop two individuals $p_1$ and $p_2$ are selected uniformly at random to serve as parents and two tours are generated from these parents; one by EAX-1AB~CO ($p_3$; focus on solution quality) and another with EAX-EDO~CO ($p_4$; focus on diversity). $p_1$ is replaced with $p_3$ if $p_3$ has lower costs than $Best$ or it has a  lower cost than $p_1$ and the algorithm has not violated the $M$ consecutive fitness evaluations without improvement in $Best$. Otherwise, $p_4$ is added to the population if $c(p_4) \leq c_{max}$. Next, if the size of the population is $\mu + 1$, the algorithm drops the individual $p \in P \setminus P^*$ whose deletion results in the least decrease in population diversity.
A subset $P^*$ of the best $k\%$ 
of the population in terms of costs always remains in the population to avoid loss of high quality candidates until $M$ consecutive failures in improvement of $Best$. Eventually, $P^*$ and $c_{max}$ are updated and the next iteration begins. These steps continue until a termination criterion is met. Note that it is not required to generate the two offspring separately. The vast majority of necessary calculations are identical for both tours. Thus, one can simultaneously generate both tours with single calculation to decrease computational costs.     

\begin{algorithm}[t!]
\begin{algorithmic}[1]
\REQUIRE{Initial Population $P$, and a limit for consecutive failures in improvement of the shortest tour $M$.}
\STATE Store the $k\%$ of $P$ with shortest tours in $P^*$.
\STATE Store the shortest tour in $Best$
\STATE Let $c_{max}$ be the maximum cost within the population.
\STATE $q \gets 0$ \COMM{counts the number of consecutive failures in improvement of the shortest tour}
\WHILE{termination criterion is not met}
\STATE Choose $p_1$ and $p_2 \in P$ based on parent selection procedure, produce two offspring $p_3$ and $p_4$ by Crossover.\\ 
\IF{$c(p_3) < Best$}
\STATE Replace $p_1$ with $p_3$ in $P$ and update $P^*$ and $Best$, and set $q = 0$
\ELSIF{$c(p_3) < c(p_1)$ \& $q < M$}
\STATE Replace $p_1$ with $p_3$ in $P$, update $P^*$, and set $q \gets q+1$.
\ELSIF{$c(p_4) \leq c_{max}$}
\STATE Add $p_4$ to $P$.
\IF{q < M}
\STATE Remove one individual $p$ from $P$, where $p = \argmax_{q \in P\setminus P^*} H(P \setminus \{q\})$ and update $c_{max}$
\ELSE
\STATE Remove one individual $p$ from $P$, where $p = \argmax_{q \in P\setminus Best} H(P \setminus \{q\})$ and update $c_{max}$
\ENDIF
\STATE $q \gets q+1$
\ELSE
\STATE Set $q \gets q+1$
\ENDIF
\ENDWHILE
\end{algorithmic}
\caption{Single-stage EAX-EDO}
\label{alg:single_stage}
\end{algorithm}

\begin{table*}[ht]
\centering
\caption{Comparison of diversity measures. In columns stat the notation $X^+$ means the median of the measure is better than the one for variant $X$, $X^-$ means it is worse and $X^*$ indicates no significant difference. Stat indicates the results of Kruskal-Wallis statistical test at significance level $5\%$ and Bonferroni correction.}
\label{tab:Known}
\renewcommand{\tabcolsep}{2pt}
\renewcommand{\arraystretch}{1.6}
\begin{small}
\begin{tabular}{lrrccccccccccccccccccc}
\toprule
& & & \multicolumn{6}{c}{\textbf{ENT} (1)} & 
 \multicolumn{6}{c}{\textbf{ED} (2)} &
 \multicolumn{6}{c}{\textbf{PD} (3)} \\
 \cmidrule(l{2pt}r{2pt}){4-9}
 \cmidrule(l{2pt}r{2pt}){10-15}
 \cmidrule(l{2pt}r{2pt}){16-21}
& $\mu$ & $\alpha$ & \textbf{H} & stat & \textbf{ED} & stat & \textbf{PD} & stat & \textbf{H} & stat & \textbf{ED} & stat & \textbf{PD}  & stat & \textbf{H} & stat & \textbf{ED} &stat &  \textbf{PD} & stat \\
\midrule
eil51&50&0.05&\hl{\textbf{0.60}}&$2^+3^+$&\hl{\textbf{0.35}}&$2^*3^+$&0.05&$2^*3^-$&0.51&$1^-3^-$&0.32&$1^*3^+$&0.02&$1^-3^-$&0.52&$1^-2^+$&0.28&$1^-2^-$&\hl{\textbf{0.18}}&$1^+2^+$\\
eil51&50&0.1&\hl{\textbf{0.86}}&$2^+3^+$&\hl{\textbf{0.47}}&$2^*3^+$&0.12&$2^+3^-$&0.73&$1^-3^-$&0.46&$1^*3^+$&0.04&$1^-3^-$&0.79&$1^-2^+$&0.41&$1^-2^-$&\hl{\textbf{0.30}}&$1^+2^+$\\
eil51&50&0.5&\hl{\textbf{1.79}}&$2^+3^+$&0.77&$2^-3^*$&0.55&$2^+3^-$&1.60&$1^-3^-$&\hl{\textbf{0.81}}&$1^+3^+$&0.20&$1^-3^-$&1.75&$1^-2^+$&0.78&$1^*2^-$&\hl{\textbf{0.71}}&$1^+2^+$\\
\midrule
eil51&100&0.05&\hl{\textbf{0.60}}&$2^+3^+$&\hl{\textbf{0.34}}&$2^+3^+$&0.05&$2^*3^-$&0.50&$1^-3^-$&0.29&$1^-3^+$&0.02&$1^-3^-$&0.51&$1^-2^+$&0.26&$1^-2^-$&\hl{\textbf{0.15}}&$1^+2^+$\\
eil51&100&0.1&\hl{\textbf{0.88}}&$2^+3^+$&\hl{\textbf{0.47}}&$2^+3^+$&0.08&$2^+3^-$&0.73&$1^-3^-$&0.44&$1^-3^+$&0.03&$1^-3^-$&0.76&$1^-2^+$&0.37&$1^-2^-$&\hl{\textbf{0.24}}&$1^+2^+$\\
eil51&100&0.5&\hl{\textbf{1.81}}&$2^+3^+$&0.77&$2^-3^+$&0.39&$2^+3^-$&1.59&$1^-3^-$&\hl{\textbf{0.80}}&$1^+3^+$&0.10&$1^-3^-$&1.77&$1^-2^+$&0.77&$1^-2^-$&\hl{\textbf{0.65}}&$1^+2^+$\\
\midrule
eil76&50&0.05&\hl{\textbf{0.51}}&$2^+3^+$&\hl{\textbf{0.28}}&$2^+3^+$&0.06&$2^*3^-$&0.43&$1^-3^-$&0.24&$1^-3^+$&0.03&$1^*3^-$&0.43&$1^-2^+$&0.22&$1^-2^-$&\hl{\textbf{0.15}}&$1^+2^+$\\
eil76&50&0.1&\hl{\textbf{0.77}}&$2^+3^+$&\hl{\textbf{0.41}}&$2^+3^+$&0.11&$2^+3^-$&0.68&$1^-3^-$&0.38&$1^-3^+$&0.05&$1^-3^-$&0.68&$1^-2^+$&0.33&$1^-2^-$&\hl{\textbf{0.25}}&$1^+2^+$\\
eil76&50&0.5&\hl{\textbf{1.78}}&$2^+3^+$&0.75&$2^-3^+$&0.55&$2^+3^-$&1.62&$1^-3^-$&\hl{\textbf{0.8}}&$1^+3^+$&0.13&$1^-3^-$&1.74&$1^-2^+$&0.75&$1^*2^-$&\hl{\textbf{0.68}}&$1^+2^+$\\
\midrule
eil76&100&0.05&\hl{\textbf{0.50}}&$2^+3^+$&\hl{\textbf{0.26}}&$2^+3^+$&0.05&$2^*3^-$&0.41&$1^-3^-$&0.22&$1^-3^+$&0.03&$1^-3^-$&0.41&$1^-2^+$&0.19&$1^-2^-$&\hl{\textbf{0.11}}&$1^+2^+$\\
eil76&100&0.1&\hl{\textbf{0.76}}&$2^+3^+$&\hl{\textbf{0.38}}&$2^+3^+$&0.08&$2^*3^-$&0.65&$1^-3^-$&0.34&$1^-3^+$&0.04&$1^-3^-$&0.64&$1^-2^+$&0.29&$1^-2^-$&\hl{\textbf{0.19}}&$1^+2^+$\\
eil76&100&0.5&\hl{\textbf{1.79}}&$2^+3^+$&0.74&$2^-3^+$&0.35&$2^+3^-$&1.63&$1^-3^-$&\hl{\textbf{0.79}}&$1^+3^+$&0.07&$1^-3^-$&1.72&$1^-2^+$&0.69&$1^-2^-$&\hl{\textbf{0.56}}&$1^+2^+$\\
\midrule
eil101&50&0.05&\hl{\textbf{0.52}}&$2^+3^+$&\hl{\textbf{0.28}}&$2^+3^+$&0.07&$2^*3^-$&0.45&$1^-3^-$&0.24&$1^-3^+$&0.05&$1^*3^-$&0.43&$1^-2^+$&0.21&$1^-2^-$&\hl{\textbf{0.15}}&$1^+2^+$\\
eil101&50&0.1&\hl{\textbf{0.75}}&$2^+3^+$&\hl{\textbf{0.39}}&$2^+3^+$&0.11&$2^+3^-$&0.67&$1^-3^-$&0.37&$1^*3^+$&0.05&$1^-3^-$&0.66&$1^-2^+$&0.31&$1^-2^-$&\hl{\textbf{0.23}}&$1^+2^+$\\
eil101&50&0.5&\hl{\textbf{1.76}}&$2^+3^+$&0.72&$2^-3^+$&0.53&$2^+3^-$&1.56&$1^-3^-$&\hl{\textbf{0.76}}&$1^+3^+$&0.08&$1^-3^-$&1.71&$1^-2^+$&0.71&$1^-2^-$&\hl{\textbf{0.61}}&$1^+2^+$\\
\midrule
eil101&100&0.05&\hl{\textbf{0.49}}&$2^+3^+$&\hl{\textbf{0.24}}&$2^+3^+$&0.05&$2^*3^-$&0.42&$1^-3^-$&0.21&$1^*3^+$&0.03&$1^*3^-$&0.4&$1^-2^+$&0.18&$1^-2^-$&\hl{\textbf{0.11}}&$1^+2^+$\\
eil101&100&0.1&\hl{\textbf{0.72}}&$2^+3^+$&\hl{\textbf{0.35}}&$2^+3^+$&0.06&$2^*3^-$&0.62&$1^-3^-$&0.32&$1^-3^+$&0.04&$1^-3^-$&0.58&$1^-2^+$&0.25&$1^-2^-$&\hl{\textbf{0.17}}&$1^+2^+$\\
eil101&100&0.5&\hl{\textbf{1.74}}&$2^+3^+$&0.70&$2^-3^+$&0.23&$2^+3^-$&1.54&$1^-3^-$&\hl{\textbf{0.74}}&$1^+3^+$&0.04&$1^-3^-$&1.61&$1^-2^+$&0.62&$1^-2^-$&\hl{\textbf{0.53}}&$1^+2^+$\\
\bottomrule
\end{tabular}
\end{small}

\end{table*}

\section{Experimental Investigation}
\label{sec:experiments}

We perform extensive experiments in order to evaluate the introduced algorithms and the EAX-EDO CO in the settings when the optimal solution for the TSP is known and unknown.

\subsection{Known Optimal Solution}

First, we compare results where $ED$, $PD$, and $H$ are incorporated into Algorithm~\ref{alg:diversity_maximizing_EA} as fitness function in order to select the best diversity measure. Having selected the diversity measure, we examine performance of the operators EAX-EDO~CO, EAX-1AB and 2-OPT.  
%
%
For this purpose, we conduct experiments for all combinations of $\mu \in \{50,100\}$ and $\alpha \in \{0.05,0.1,0.5\}$ on instances eil51, eil76, and eil101 from the TSPlib~\cite{Reinelt91tsplib} for $10$ independent runs.

\paragraph{Comparison between diversity measures}
In this subsection, we examine the performance of  Algorithm~\ref{alg:diversity_maximizing_EA} in the case where $H$, $ED$, and $PD$ is embedded into the algorithm as the fitness function. The algorithm is initialised with $\mu$ copies of an optimal solution, and 2-OPT is used as the operator to generate offspring.   

Table~\ref{tab:Known} shows that the algorithm using entropy $H$ as the fitness function not only outperforms its counterparts in terms of entropy value ($H$) in all the cases. It also leads to higher $ED$ scores compared to the algorithm using $ED$ as fitness function when $\alpha$ is equal to $0.05$ or $0.1$. Moreover, the entropy-based algorithm ($H$) results in higher $PD$ scores compared to the $ED$-based algorithm, and higher $ED$ score in comparison to the $PD$-based algorithm. These observations are supported by the results of a Kruskal-Wallis test at significance level $5\%$ and Bonferroni correction. Therefore, the entropy-based measure is selected as the fitness function for the following experiments. 

\paragraph{Comparison between EAX-EDO CO, EAX-1AB CO, 2-OPT}
Here, we investigate the performance of the proposed EAX-EDO~CO in comparison to EAX-1AB and 2-OPT. Note that we used 2-OPT in the first $1\,000$ iterations for all competitors since EAX-EDO~CO and EAX-1AB as crossover-operators require two different parents to generate an offspring that is no clone, while the initial population consists of $\mu$ copies of a single solution. The experimental settings and instances are in line with the previous subsection.
\begin{figure}[ht]
    \centering
    \includegraphics[width=1\columnwidth,scale=1,trim=0 0pt 0 0, clip]{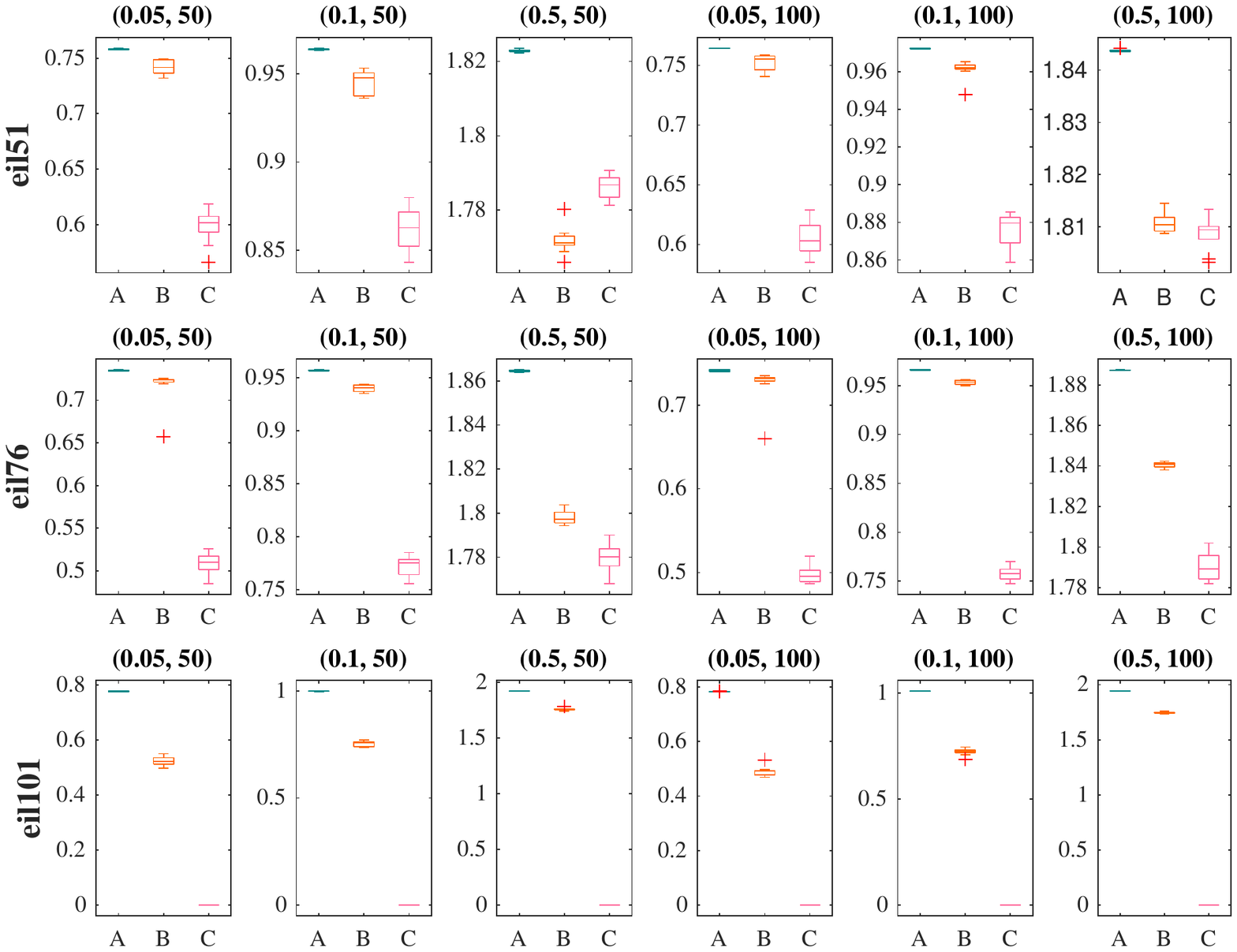} 
    \caption{Distributions of diversity scores (to be maximised) of populations obtained by Algorithm~\ref{alg:diversity_maximizing_EA} with EAX-EDO CO~(A), EAX-1AB CO~(B), and 2-OPT~(C) for instances eil51, eil76, and eil101. Labels above the box-plots indicate $(\alpha, \mu)$.}
    \label{fig:EDO-1AB}
\end{figure}

Figure~\ref{fig:EDO-1AB} shows the performance of EAX-EDO~CO, EAX-1AB~CO, and 2-OPT (encapsulated in Algorithm~\ref{alg:diversity_maximizing_EA}). The figure illustrates that not only EAX-EDO~CO outperforms EAX-1AB~CO and 2-OPT in terms of the mean diversity score in all cases but also has lower standard deviation. These observations are confirmed by the results of a Kruskal-Wallis test at significance level $5\%$ and Bonferroni correction that indicates a significant difference is found in the median of entropy scores of the final populations obtained by EAX-EDO~CO and that of the EAX-1AB and 2-OPT in all considered settings.
Turning to the comparison between EAX-1AB and 2-OPT, the former brings about more diverse populations in most of the cases, except for the setting with $\alpha = 0.5$ and $\mu = 50$ on eil51. One can notice that the smaller $\alpha$ is, the larger is the gap between EAX-based crossovers and 2-OPT. This is because 2-OPT is a random neighbourhood search; therefore, the tighter the threshold is, the lower the chance to generate an offspring with acceptable quality. Note that a larger neighborhood search would be necessary to escape local optima.


\subsection{Unknown Optimal Solution}
The two-stage EAX-EDO introduced for unknown optimal cases includes several parameters such as parameters associating to allocation of the budget between the cost minimisation and diversity maximisation phases. It is crucial to tune the parameters in order to deliver the algorithm's the best performance.

\subsubsection{Algorithm Configuration}

The goal of automatic algorithm configuration is to find, in an automated fashion, a parameter configuration to deliver the algorithms' best (average) performance for a given set of instances. Here, the algorithm parameters are tuned by iRace~\cite{lopez2016irace}, which performs an iterated racing procedure between different parameter configurations to find the best parameter setting. In our experimental investigation, we consider the minimum budget of $96$ runs to be performed by iRace for each algorithm due to the fact that the algorithms are computationally expensive.

The two-stage algorithm includes four input parameters. Since the budget is limited, there are three parameters associated to allocation of the budget, which include budget allocation to each repetition of the main loop ($X$), and the proportions of $X$ are allocated to cost minimisation and diversity maximisation phases, $x_c$ and $x_d$, respectively. The last parameter is the number of consecutive failures $M$ of the cost minimisation phase in finding a tour with better costs. Note that there exist dependencies between these parameters. First, $X$ is a proportion of the total budget ($It$). Second, $x_d$ can be determined as $1-x_c$. Finally, $M$ should be lower than the total number of repetitions of the main loop, i.e. $M_1<It/X$; therefore, we set $M = m \cdot (It/X)$, where $m \in (0, 1)$. 
\begin{table}
\caption{Tested parameter values during the tuning procedure.}
\label{tab:unconstrained_Cplex}
\centering
\renewcommand{\tabcolsep}{8pt}
\renewcommand{\arraystretch}{1.4}
\begin{tabular}{lccc}
\toprule
\textbf{Parameter} &$X$&$x_c$&$m$\\
\midrule
\textbf{Range} &$(0.03, 0.2)$&$(0.2, 0.8)$&$(0.1, 0.5)$\\
\textbf{Best setting} &$0.0614$&$0.4888$&$0.2413$\\
\bottomrule
\end{tabular}
\label{tab:Tuning}
\end{table}


Table~\ref{tab:Tuning} shows the parameter ranges considered in the tuning procedure and the best setting found by iRace. The other parameters can be calculated from aforementioned equations.


\subsubsection{Experiments}

We now compare the performance of single-stage and two-stage EAX-EDO against standard EAX and the Gurobi optimiser~\cite{gurobi} in terms of solution quality and entropy-based diversity. The Gurobi optimizer is a well-known mixed integer programming~(MIP) solver. Although Gurobi is usually used to obtain the optimal or a high-quality solution for a given optimisation problem, it is capable of providing its users with $\mu$ different solutions within a specific gap $\alpha$ to the optimal solution. Here, we use the Dantzig–Fulkerson–Johnson formulation~\cite{dantzig2016linear}. We use Gurobi to generate $\mu$ different solutions and use it as a baseline for comparison in our studies. 

The benchmark instances considered in this section include eil101, a280, pr493, u574, rat575, p654, rat783, u1060, pr2392, and fnl4461~\cite{Reinelt91tsplib}. Moreover, we set $\mu$ to~50.
Note that all algorithms are initialized with a population of individuals optimised by 2-OPT. The results are summarised in Table~\ref{tab:Unkown}.

\begin{table*}[t]
\centering
\caption{Comparison of the proposed algorithms with EAX in terms of diversity~($H$) and solution quality of the best solution~($c$). Here, $\Delta H$ shows the entropy of the final population $P$ on top of $H_{min}$, i.e, $\Delta H = H(P) - H_{min}$. Columns $stat_H$ and $stat_c$ contain the results of Kruskal-Wallis tests on entropy of the final population and best tour length respectively.}

\label{tab:Unkown}
\renewcommand{\tabcolsep}{2.3pt}
\renewcommand{\arraystretch}{1.6}
\begin{small}
\begin{tabular}{lcrrrrrrrrrrrrrrr}
\toprule
& & & \multicolumn{2}{c}{\textbf{Gurobi}} &  
 \multicolumn{4}{c}{\textbf{Single-stage EAX-EDO}~(1)} & 
 \multicolumn{4}{c}{\textbf{Two-stage EAX-EDO}~(2)} &
 \multicolumn{4}{c}{\textbf{EAX}~(3)}\\
 \cmidrule(l{2pt}r{2pt}){4-5}
 \cmidrule(l{2pt}r{2pt}){6-9}
 \cmidrule(l{2pt}r{2pt}){10-13}
 \cmidrule(l{2pt}r{2pt}){14-17}
& $H_{min}$ & $OPT$ & $\Delta H$ & $c$ & $\Delta H$ & $c$ & stat$_{H}$ & stat$_{c}$ & $\Delta H$ & $c$ & stat$_{H}$ & stat$_{c}$ & $\Delta H$ & $c$ & stat$_{H}$ & stat$_{c}$ \\
\midrule
eil101&5.31&629&0.11&629&\hl{\textbf{0.79}}&629&$2^+3^+$&$2^*3^*$&0.34&629&$1^-3^+$&$1^*3^*$&0.11&629&$1^-2^-$&$1^*2^*$\\
a280&6.33&$2\,579$&0.1&$2\,579$&\hl{\textbf{0.6}}&$2\,579$&$2^+3^+$&$2^*3^*$&0.30&$2\,579$&$1^-3^+$&$1^*3^*$&0.12&$2\,579$&$1^-2^-$&$1^*2^*$\\
pr439&6.78&$107\,217$&0.03&$107\,217$&\hl{\textbf{0.66}}&$107\,262$&$2^-3^*$&$2^-3^*$&0.27&$107\,217$&$1^*3^+$&$1^+3^*$&0.05&$107\,226$&$1^-2^-$&$1^*2^*$\\
u574&7.05&$36\,9054$&0.02&$36\,905$&\hl{\textbf{0.67}}&$36\,914$&$2^+3^+$&$2^-3^-$&0.29&$36\,908$&$1^-3^+$&$1^*3^*$&0.01&$36\,905$&$1^-2^-$&$1^*2^*$\\
rat575&7.05&$6\,773$&0.05&\hl{\textbf{$6\,773$}}&\hl{\textbf{0.63}}&$6\,777$&$2^+3^+$&$2^-3^-$&0.35&$6\,775$&$1^-3^+$&$1^*3^*$&0.11&$6\,774$&$1^-2^-$&$1^+2^*$\\
p654&7.18&$34\,643$&0.08&$34\,643$&\hl{\textbf{1.15}}&$34\,643$&$2^+3^+$&$2^*3^*$&0.55&$34\,646$&$1^-3^+$&$1^*3^-$&0.28&$34\,643$&$1^-2^-$&$1^*2^+$\\
rat783&7.36&$8\,806$&0.02&$8\,806$&\hl{\textbf{0.57}}&$8\,809$&$2^+3^+$&$2^*3^-$&0.31&$8\,807$&$1^-3^+$&$1^*3^*$&0.07&$8\,806$&$1^-2^-$&$1^+2^*$\\
u1060&7.66&$224\,094$&-&-&\hl{\textbf{0.69}}&$224\,275$&$2^+3^+$&$2^*3^-$&0.35&$224\,131$&$1^-3^+$&$1^*3^*$&0.07&\hl{\textbf{$224\,109$}}&$1^-2^-$&$1^+2^*$\\
pr2392&8.47&$378\,032$&-&-&\hl{\textbf{0.56}}&$378\,813$&$2^+3^+$&$2^*3^-$&0.28&$378\,926$&$1^-3^+$&$1^*3^-$&0.02&\hl{\textbf{$378\,059$}}&$1^-2^-$&$1^+2^+$\\
fnl4461&9.1&$182\,566$&-&-&0.33&\hl{\textbf{$182\,297$}}&$2^*3^-$&$2^*3^+$&0.32&$183\,200$&$1^+3^-$&$1^-3^+$&\hl{\textbf{0.42}}&$184\,230$&$1^+2^+$&$1^-2^-$\\
\bottomrule
\end{tabular}
\end{small}

\end{table*}
\begin{figure*}
\centering
\includegraphics[width=0.24\columnwidth]{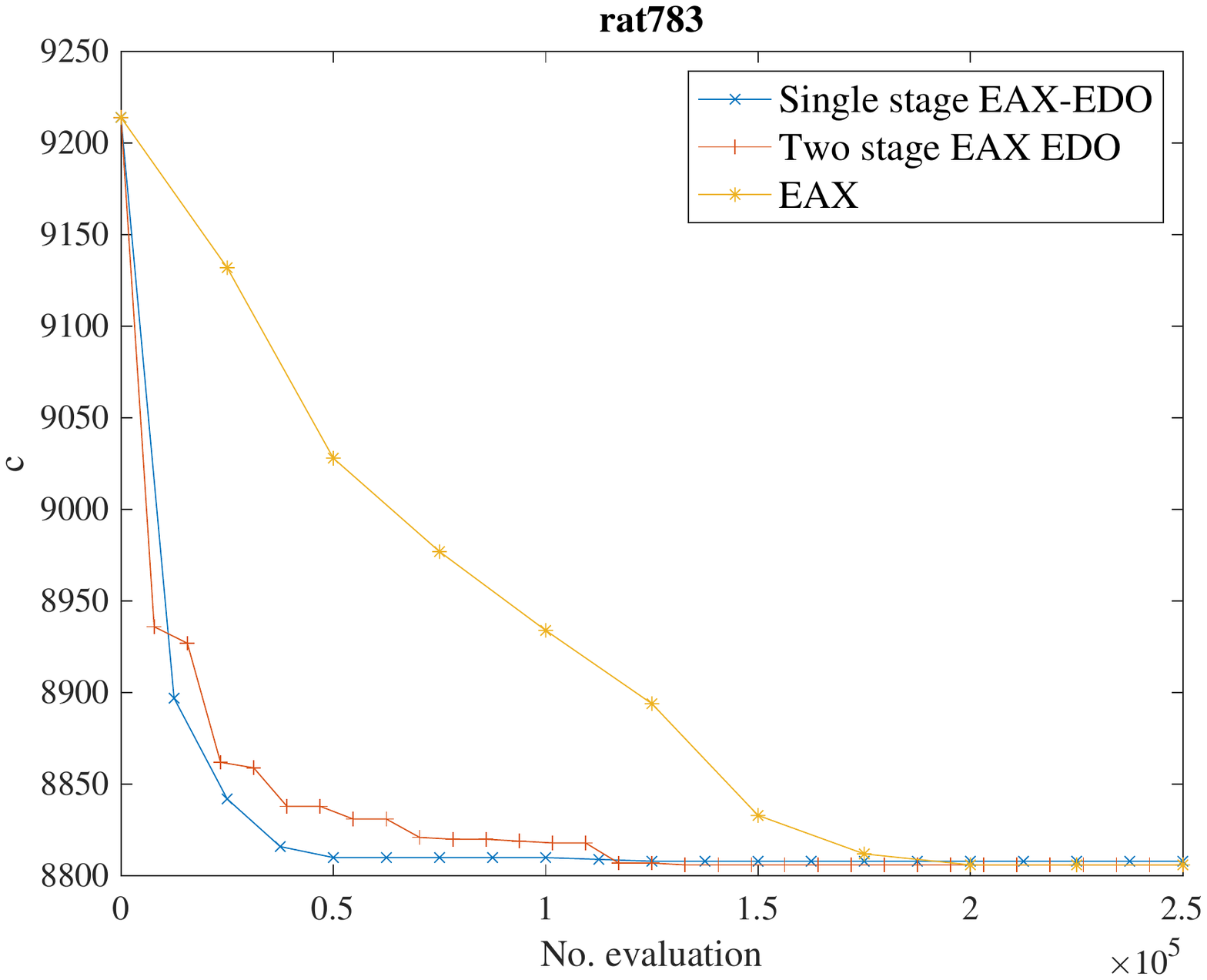}
\includegraphics[width=0.23\columnwidth]{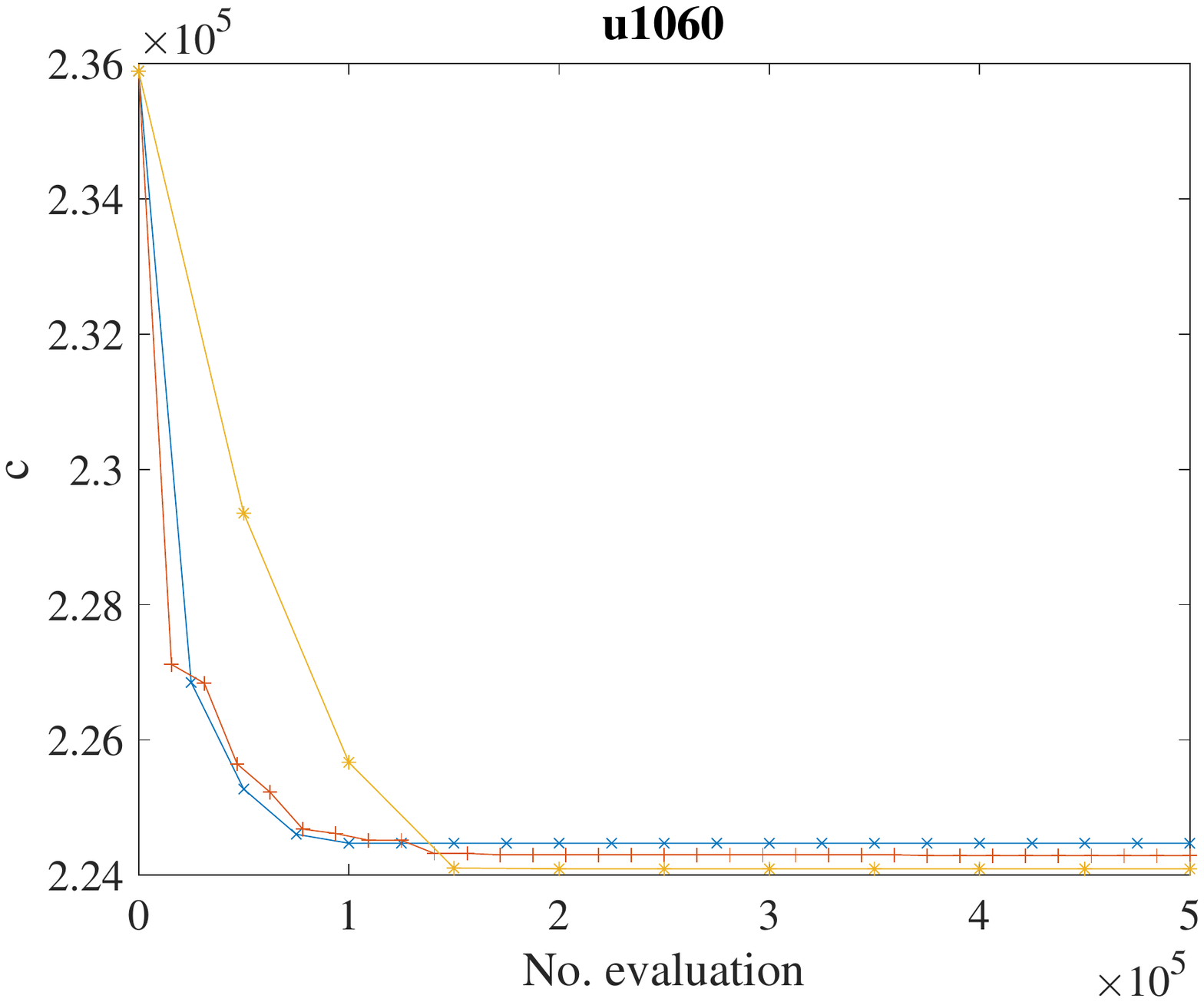}
\includegraphics[width=0.23\columnwidth]{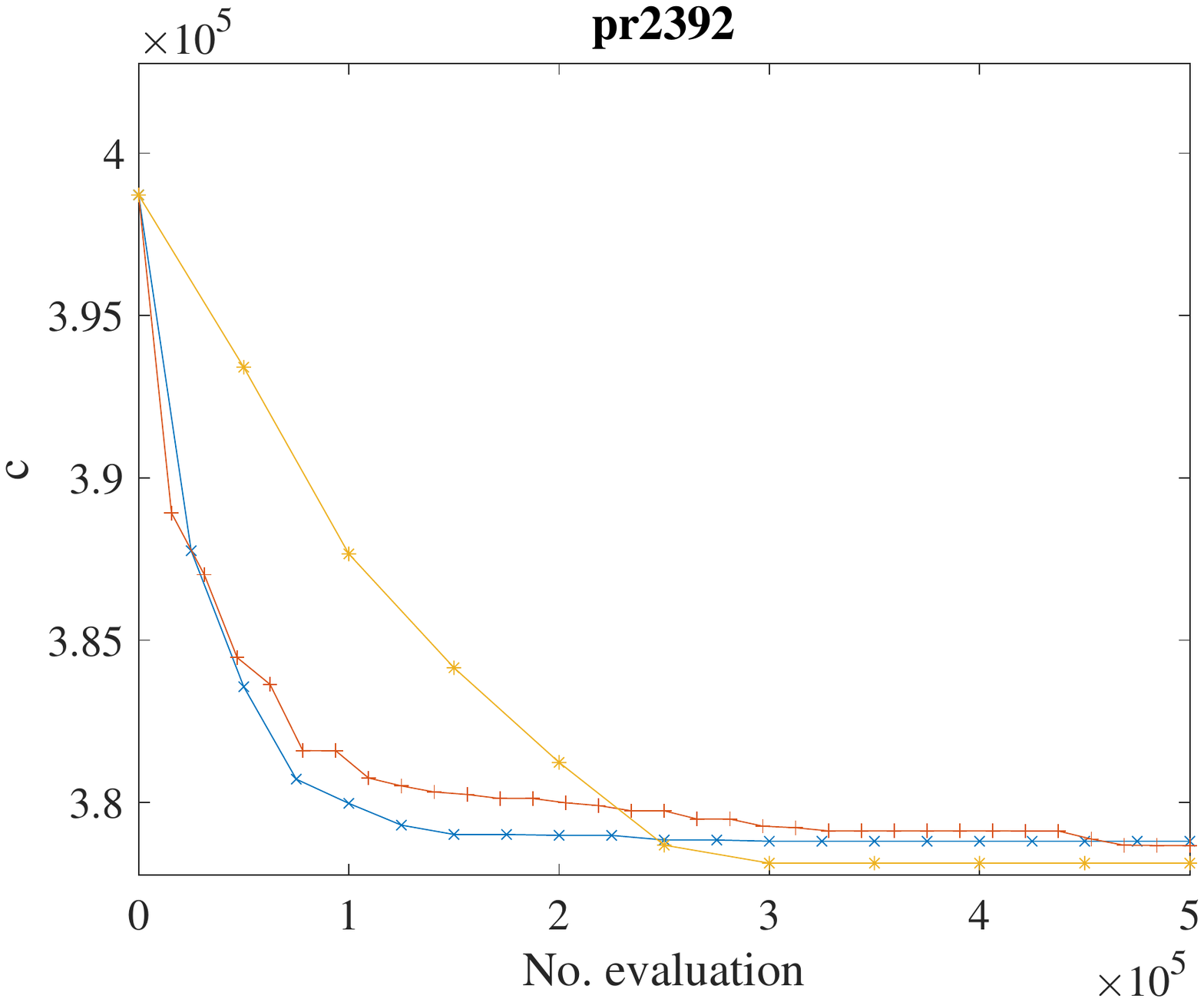}
\includegraphics[width=0.23\columnwidth]{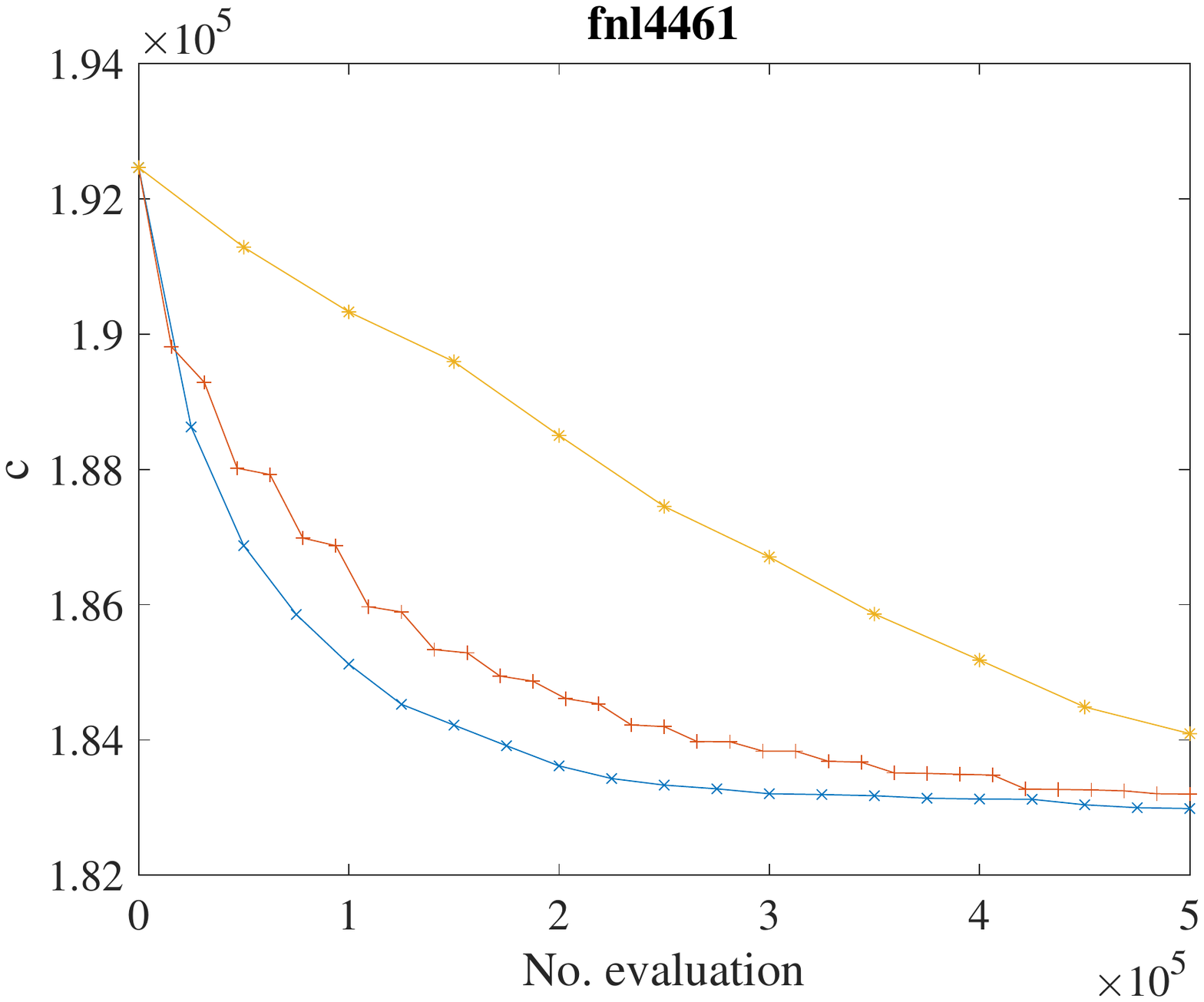}
\includegraphics[width=0.23\columnwidth]{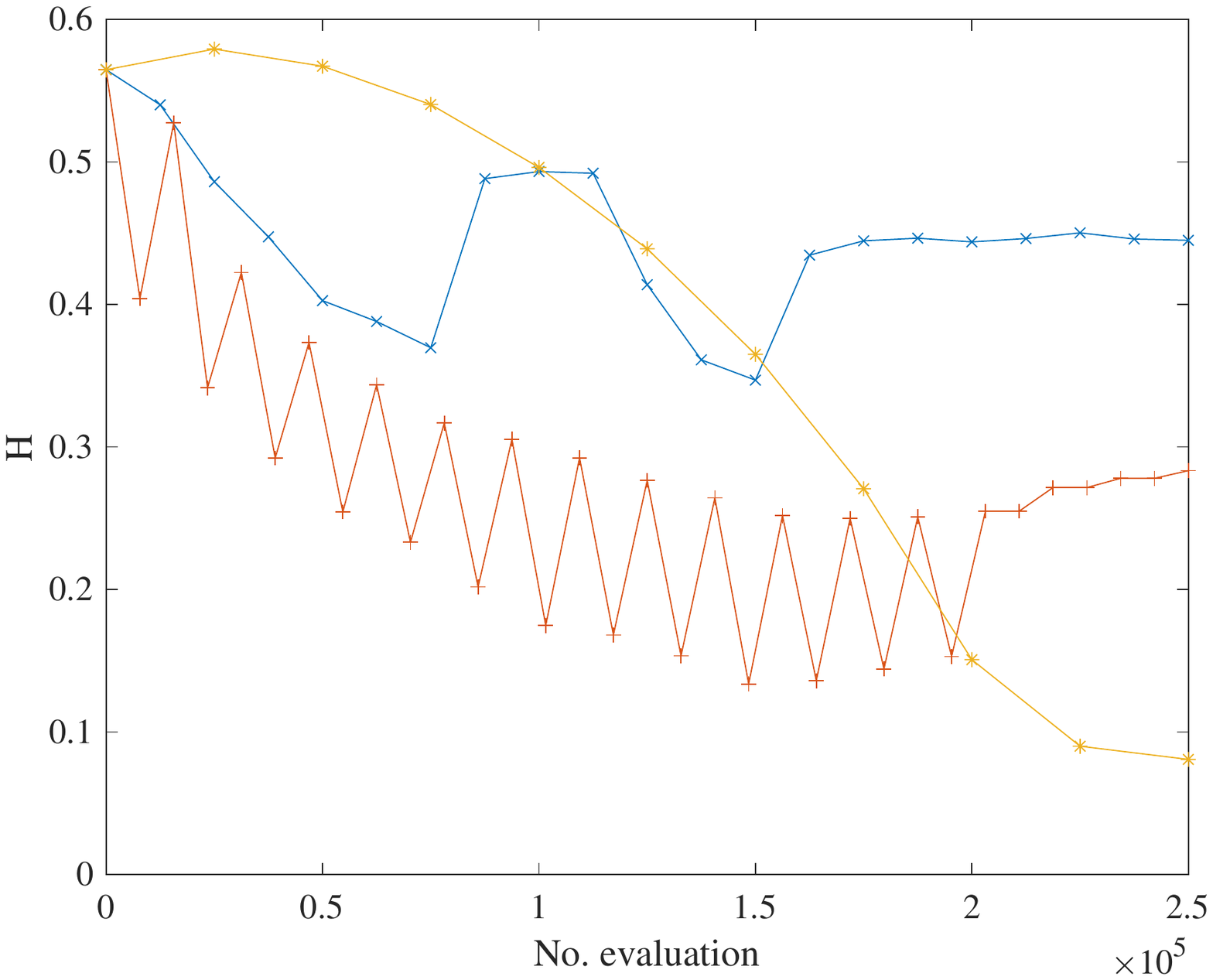}
\includegraphics[width=0.23\columnwidth]{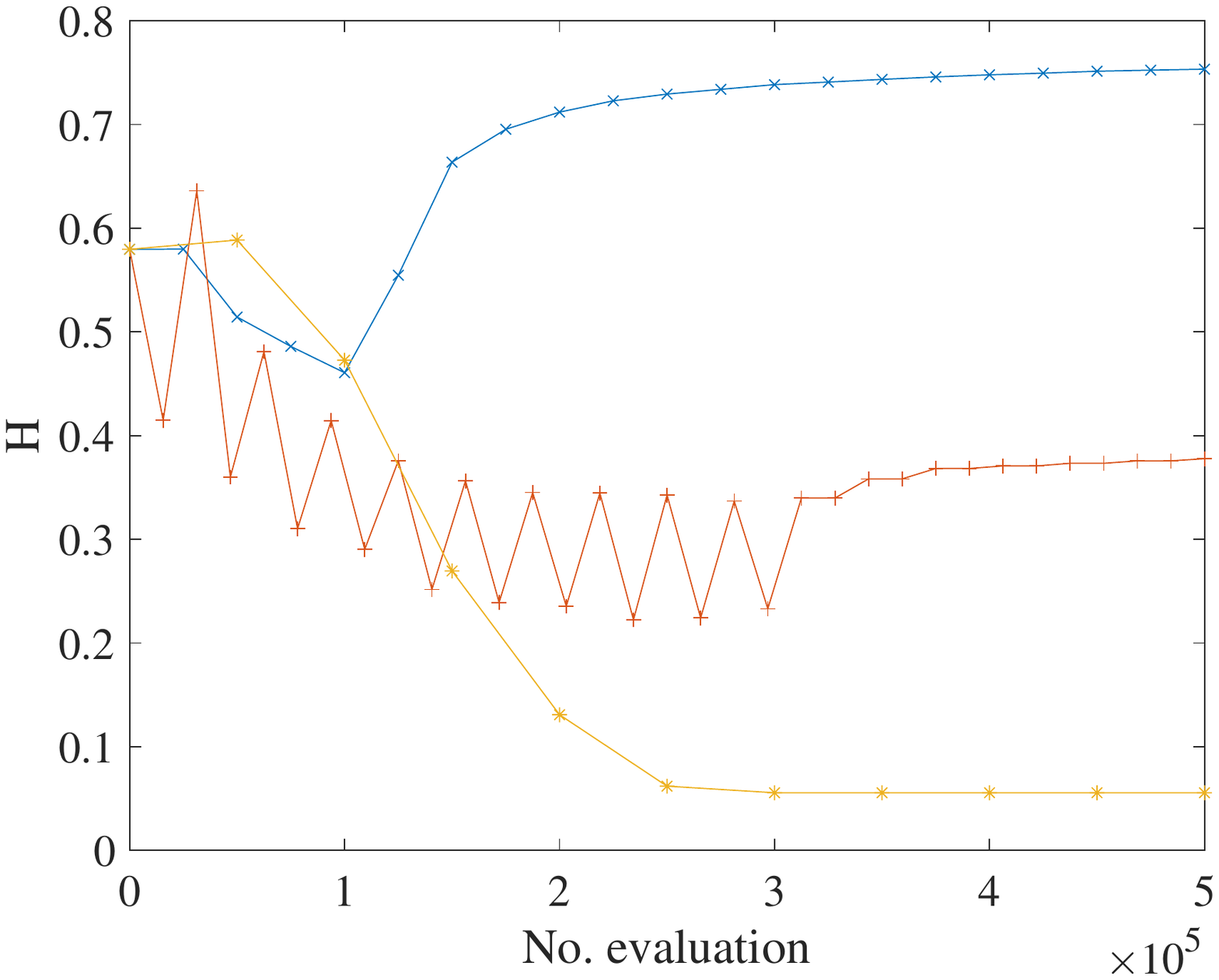}
\includegraphics[width=0.23\columnwidth]{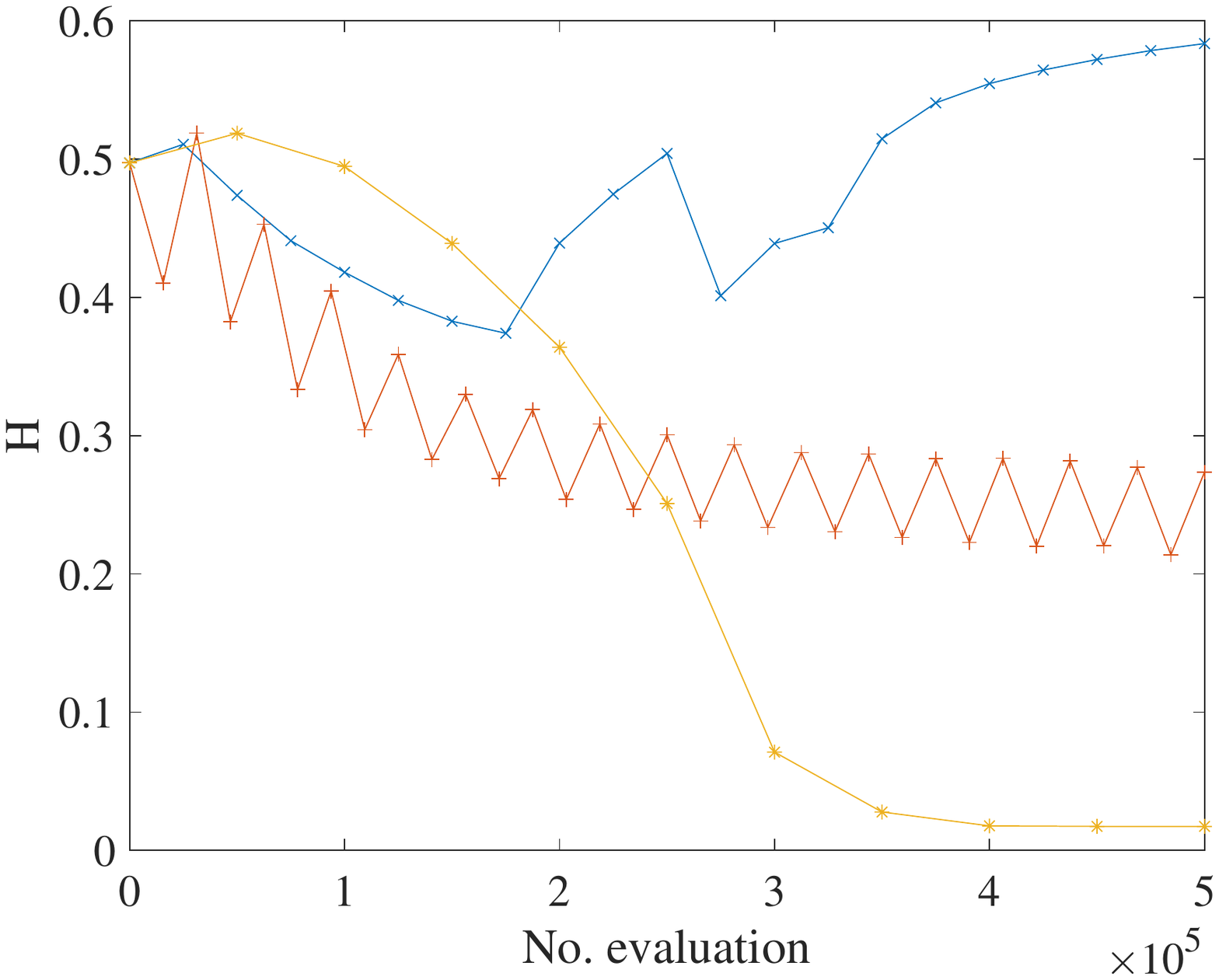}
\includegraphics[width=0.23\columnwidth]{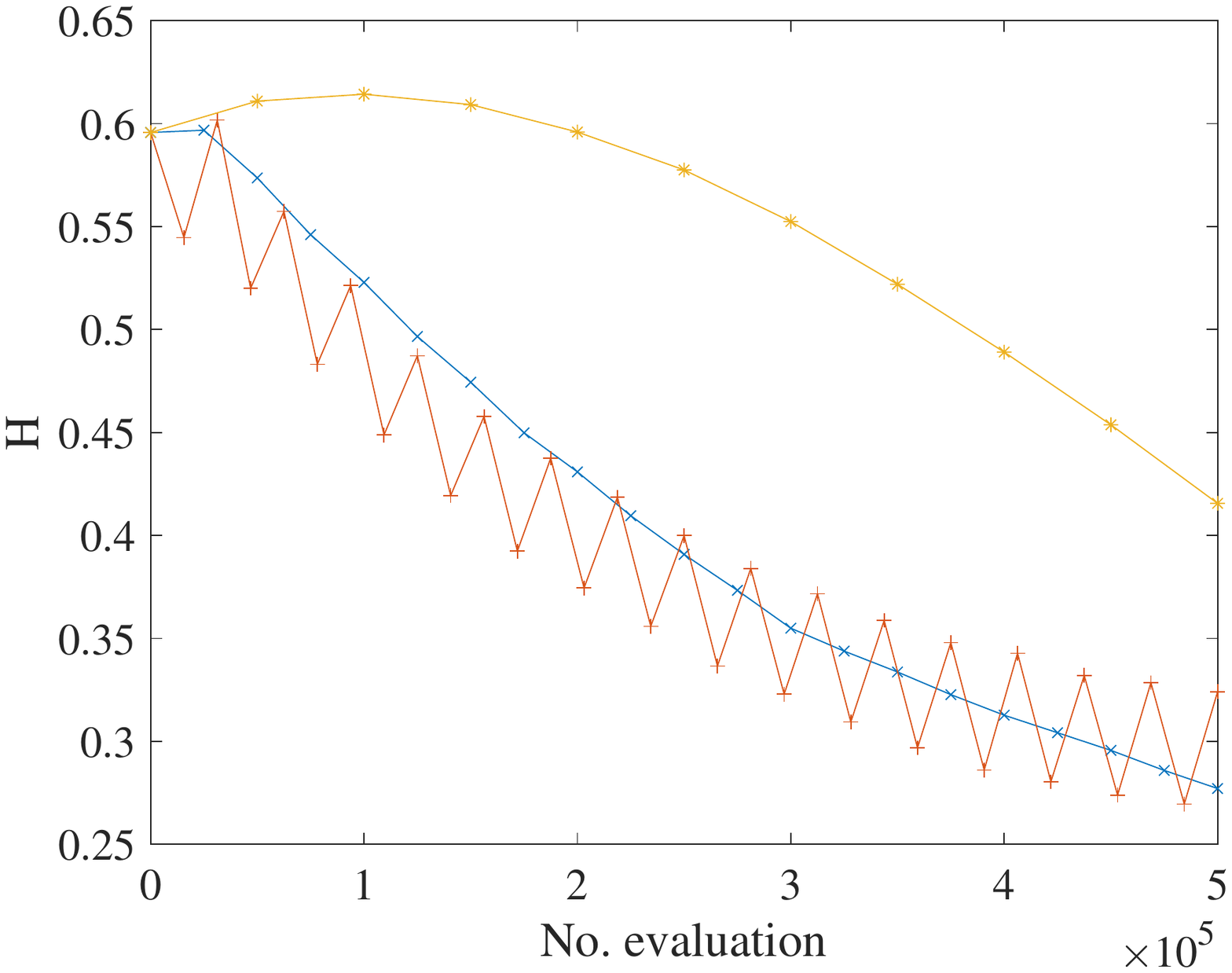}
\caption{Representative trajectories in the setting of unknown optimal solutions. The plots show the best tour length in the population (first row) and diversity measured by the entropy (second row).}
\label{fig:con_rat}
\end{figure*}
The outcome indicates that the introduced single-stage EAX-EDO outperforms the other algorithms in terms of diversity. Table~\ref{tab:Unkown} shows that both single-stage EAX-EDO and two-stage EAX-EDO are capable of computing tours with decent costs. The Gurobi optimiser and EAX result in marginally better quality, while they are outperformed by the single-stage EAX-EDO in terms of diversity on all instances except on instance fnl4461. On this instance, single-stage EAX-EDO leads to higher quality but less diverse population compared to EAX. 
This can be attributed to EAX generating $n_{ch}=25$ offspring per iteration what makes it converge slower than the two variants of EAX-EDO. Therefore, the EAX-EDO algorithms achieve better quality but less diverse populations in $500\,000$ fitness evaluations. Should EAX continue to run, we expect that it converges to a slightly higher quality but less diverse population compared to EAX-EDO; the same trend can be observed for smaller instances. Figure~\ref{fig:con_rat} illustrates this matter visually. The figure depicts the quality of best solutions and diversity of populations over fitness evaluations for the single-stage EAX-EDO, the two-stage EAX-EDO, and vanilla EAX. Figure~\ref{fig:con_rat} shows that having a lower entropy in preliminary iterations, single-stage EAX-EDO finally converges to higher entropy compared to the vanilla EAX on rat783, u1060, pr2392. It is the other way around for the quality of tours; EDO-based algorithms converge faster in terms of quality, while vanilla EAX results in slightly higher quality tours. The same patterns can be observed for other instances except fnl4461. On this instance, none of algorithms converges in this setting. Here, Figure~\ref{fig:con_rat} indicates that 500\,000 fitness evaluations are insufficient to have the algorithms converged.           

Moreover, Figure~\ref{fig:con_rat} shows that EAX sacrifices the entropy of the population to gain shorter tours. On the contrary, single-stage EAX-EDO increases the entropy while in the course of optimising the solution quality. In addition, Figure~\ref{fig:con_rat} highlights the room for improvement in the EAX-EDO algorithms. First, tuning the single-stage EAX-EDO's parameters is likely to boost performance of the algorithm, although it outperforms the other counterparts in terms of diversity in the current state. Second, the EAX-EDO algorithms generate one offspring per iteration; generating $n_{ch}$ number of offspring from the same parents and incorporating the selection procedure to choose between them (same as the standard EAX), EAX-EDO is likely to achieve even higher diversity and quality. However, selection of $\mu$ tours from $\mu+n_{ch}$ tours in a way to maximise diversity is a complicated problem. For instance, there are $\binom{\mu+n_{ch}}{\mu}$ 
possible candidates for Brute-force search.
This makes the algorithm computationally more expensive.  
\begin{figure*}
\centering
\includegraphics[width=.25\columnwidth]{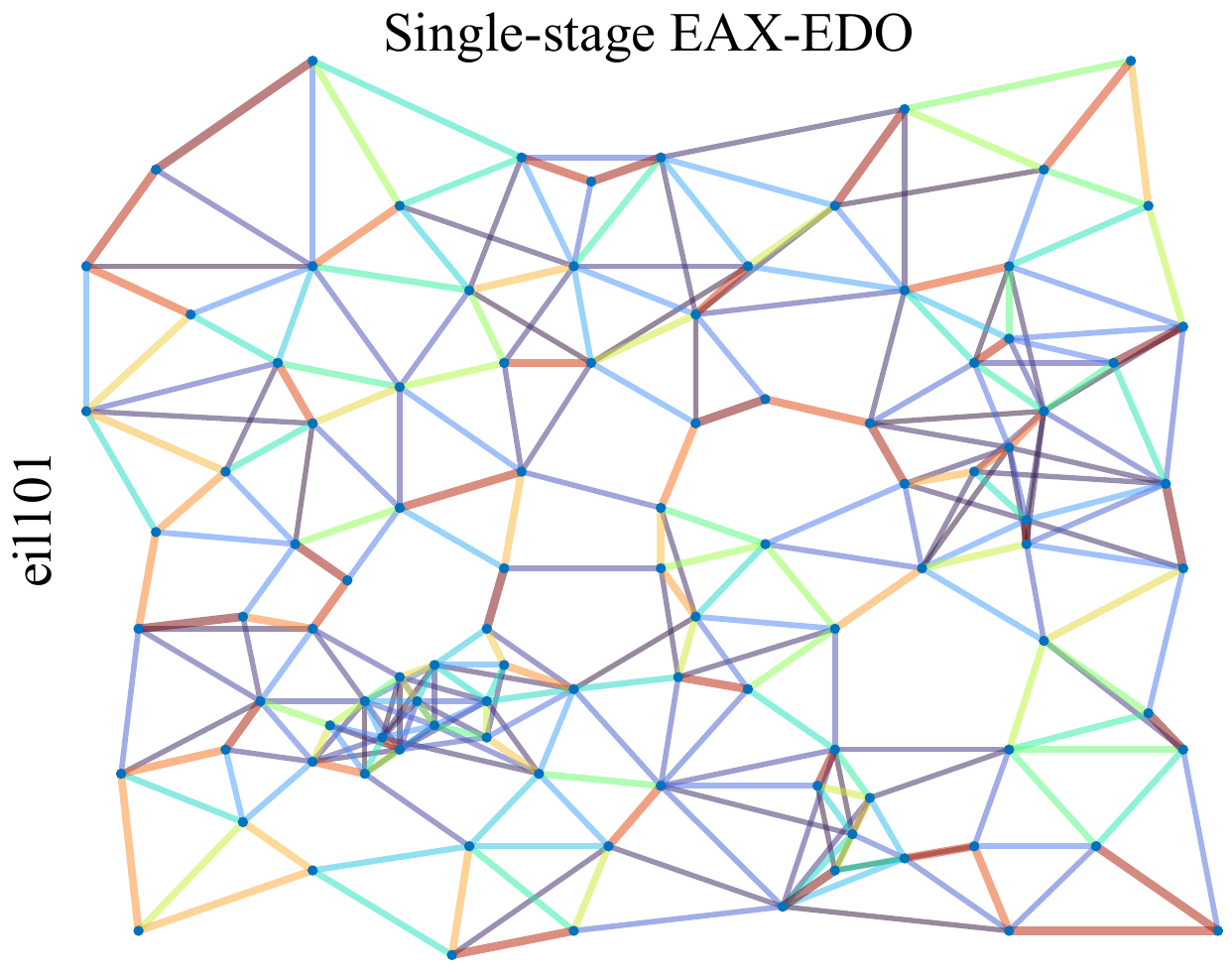}
\hskip5pt
\includegraphics[width=.23\columnwidth]{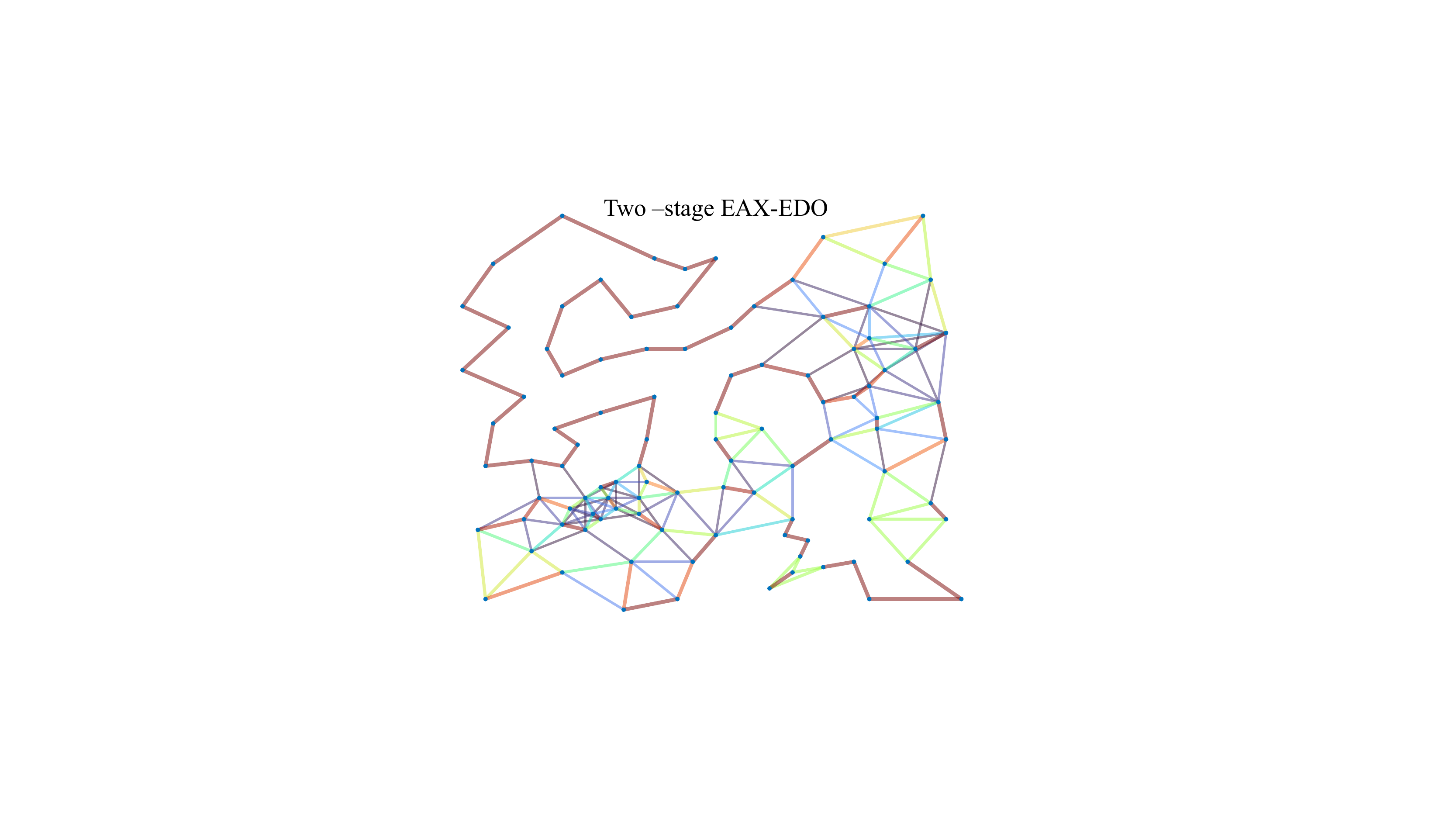}
\hskip5pt
\includegraphics[width=.23\columnwidth]{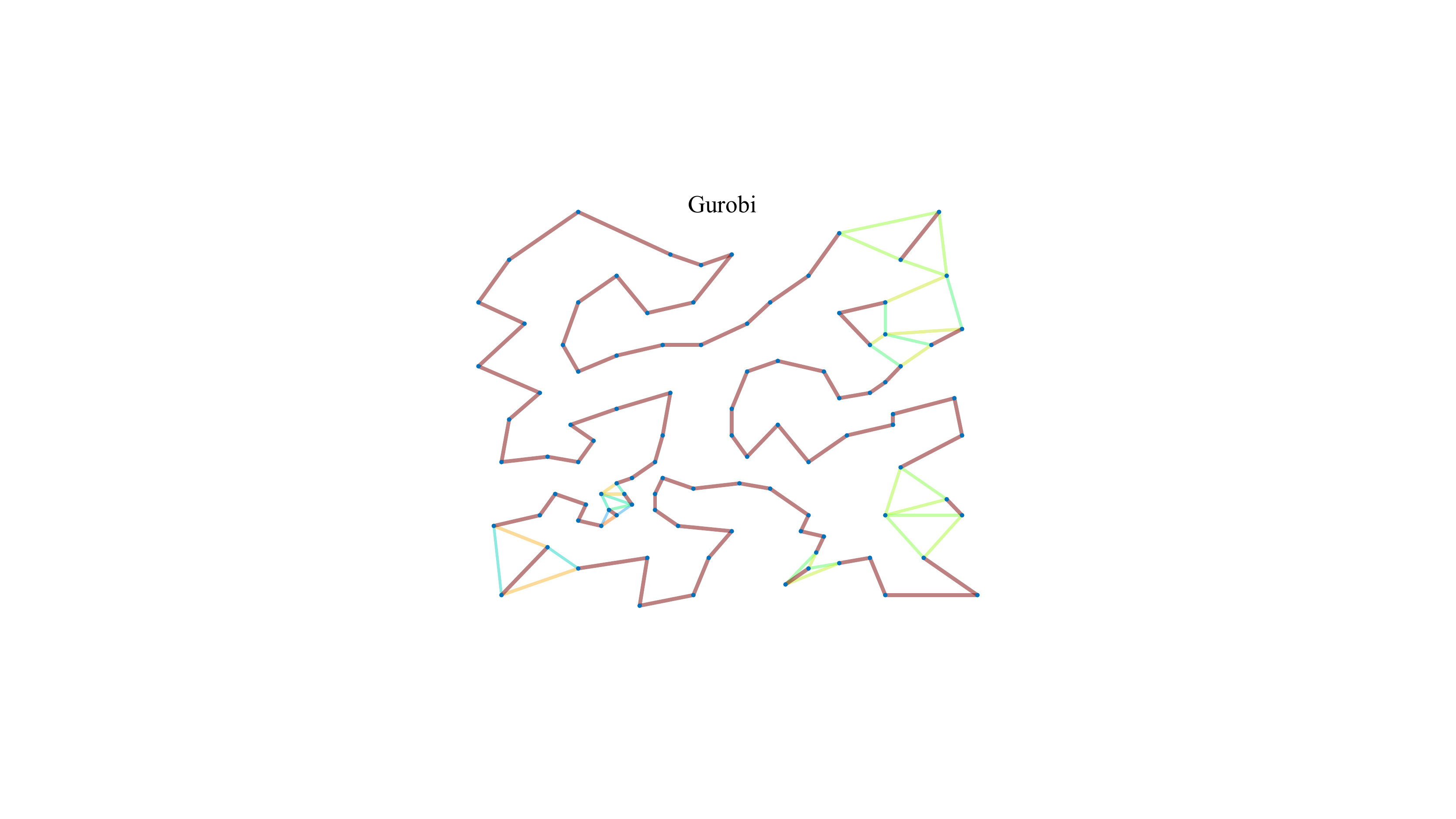}
\hskip5pt
\includegraphics[width=.23\columnwidth]{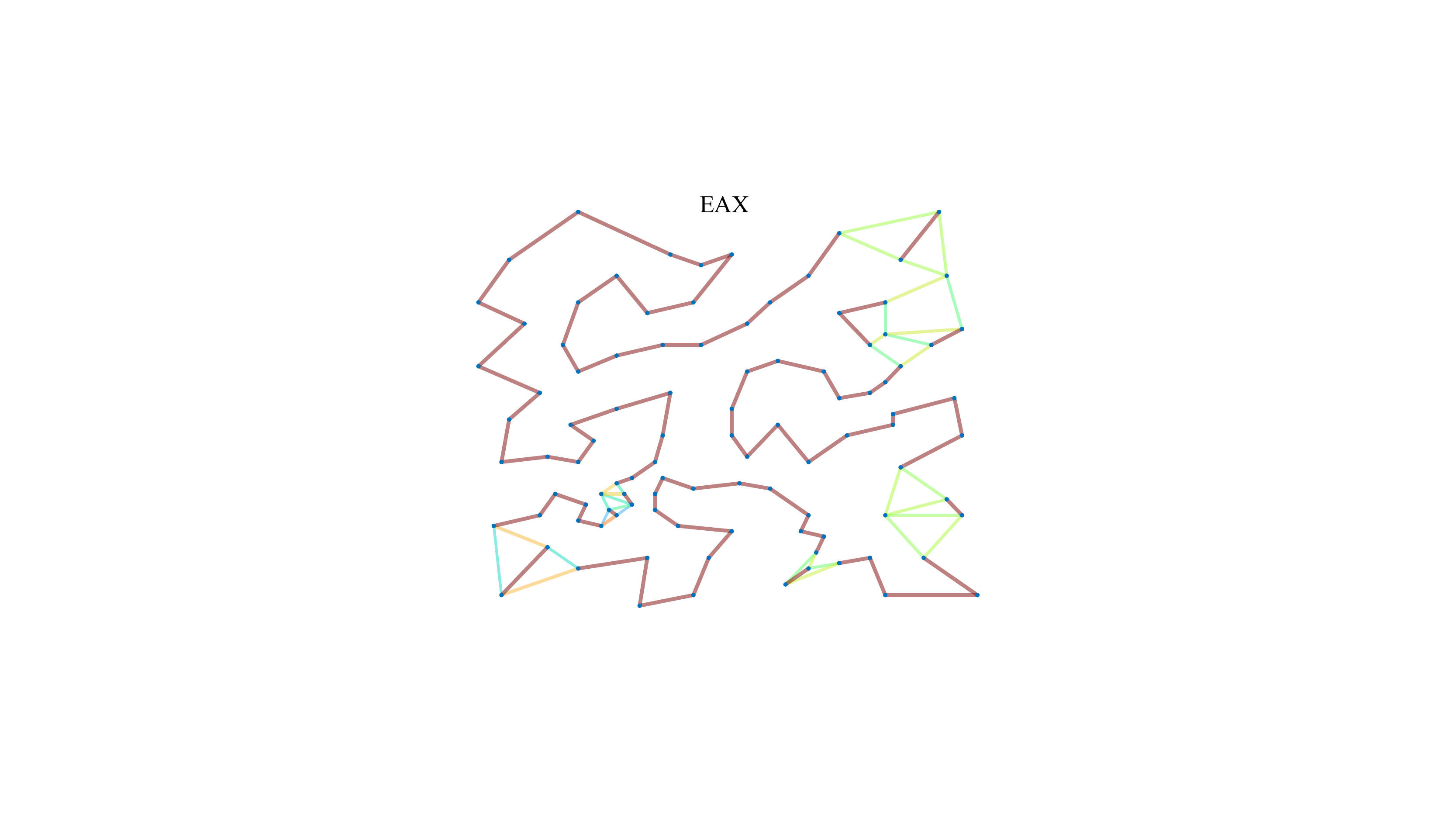}
\hskip15pt
\includegraphics[width=.23\columnwidth]{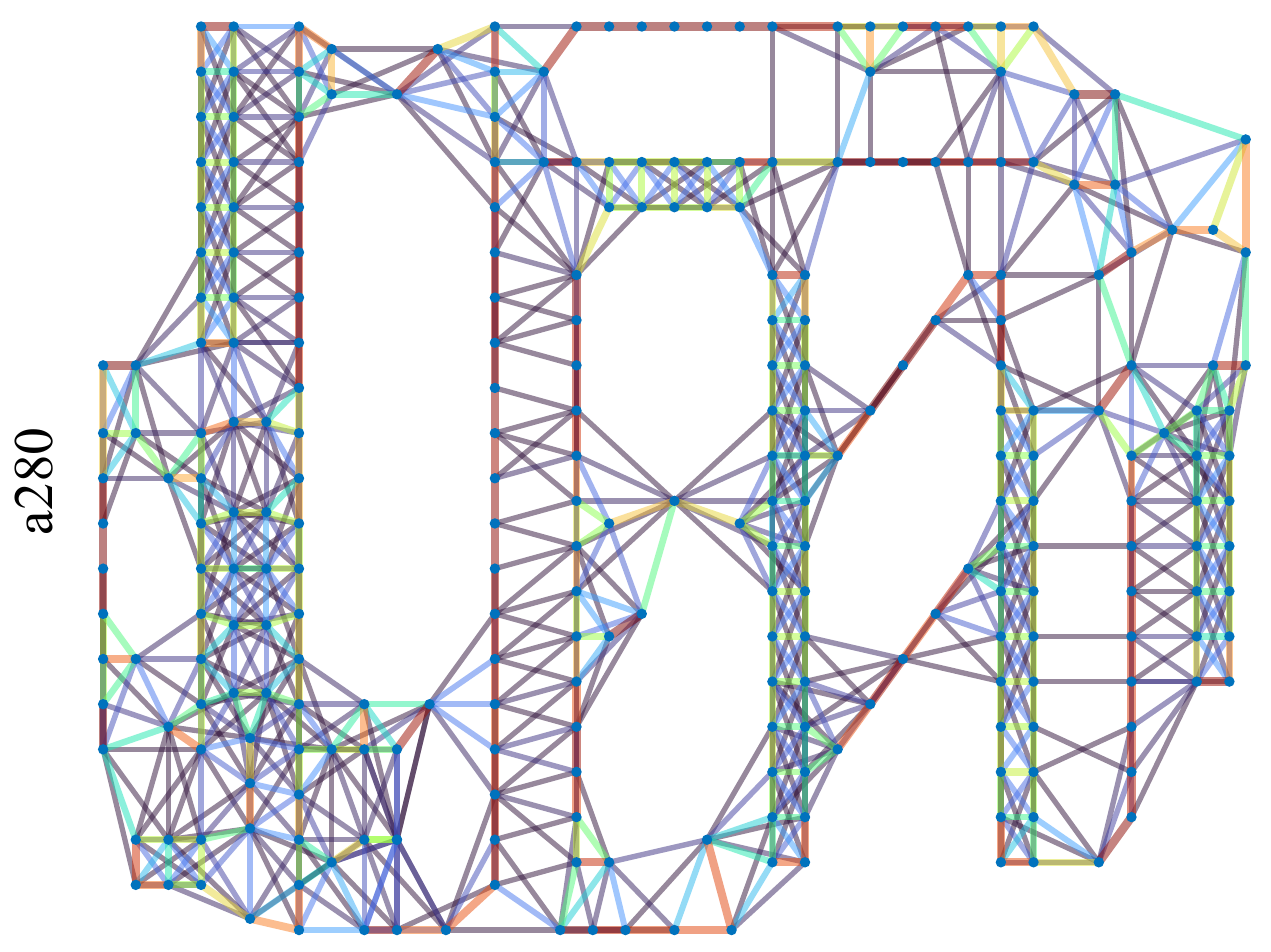}
\hskip5pt
\includegraphics[width=.23\columnwidth]{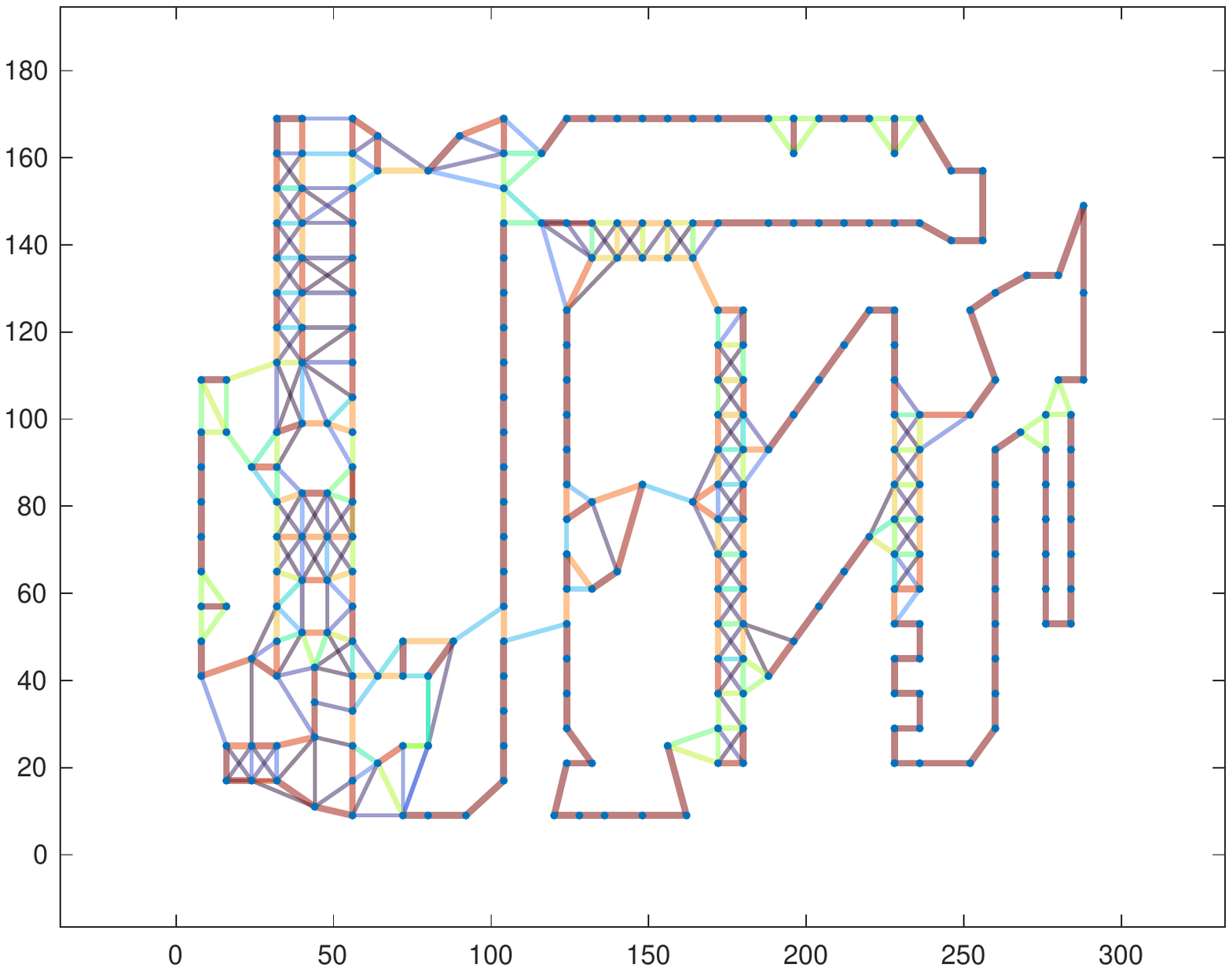}
\hskip5pt
\includegraphics[width=.23\columnwidth]{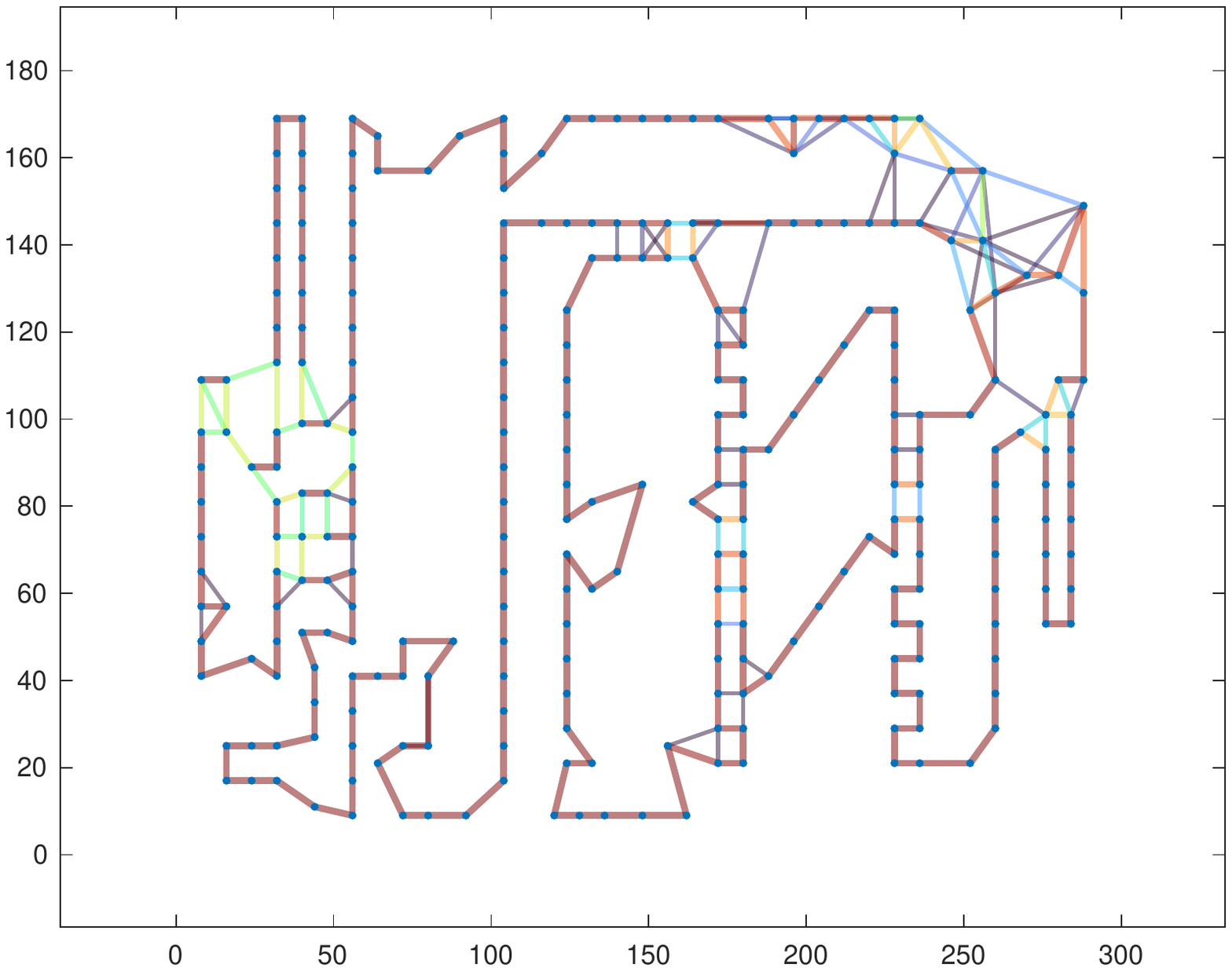}
\hskip5pt
\includegraphics[width=.23\columnwidth]{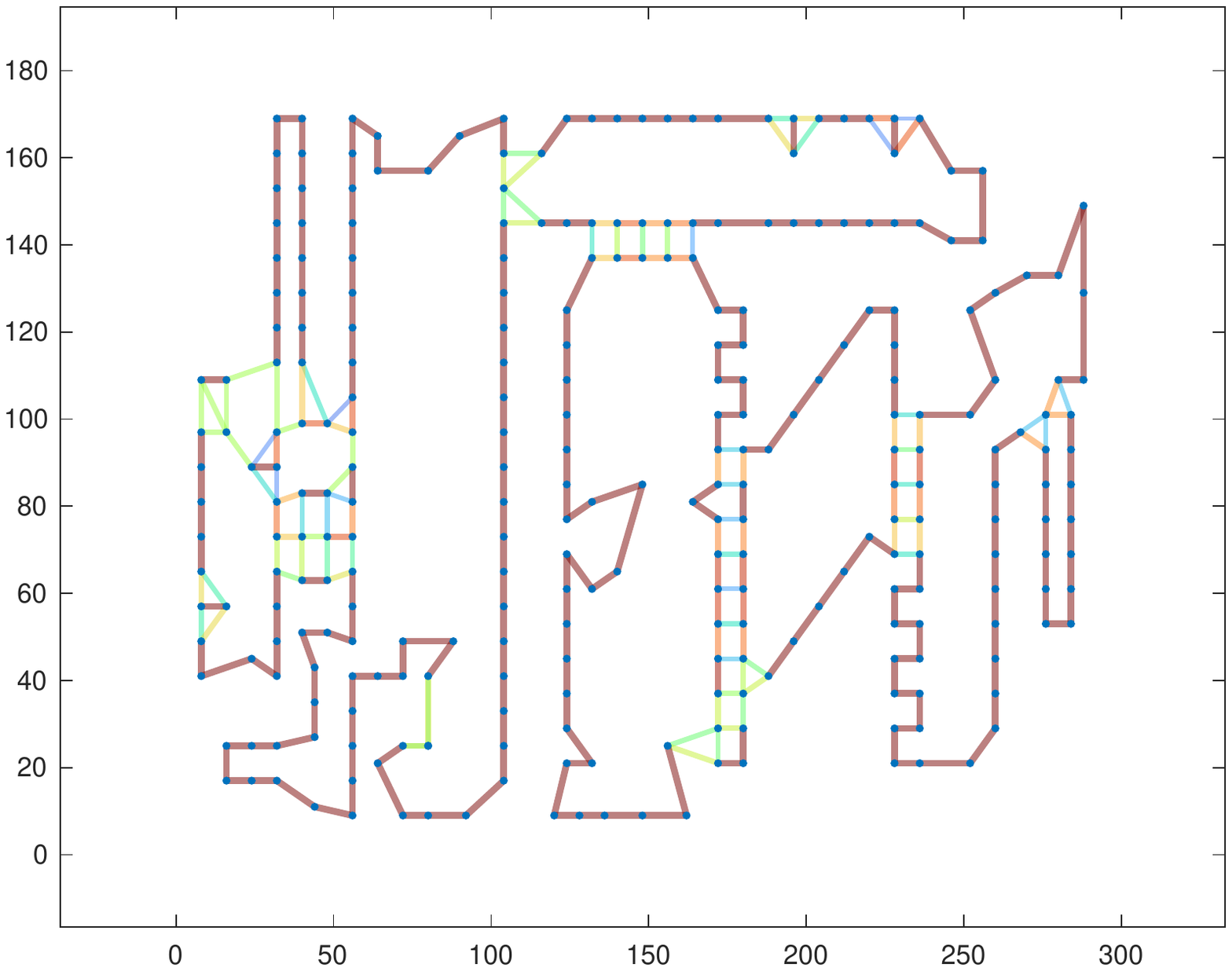}
\hskip15pt
\includegraphics[width=.23\columnwidth]{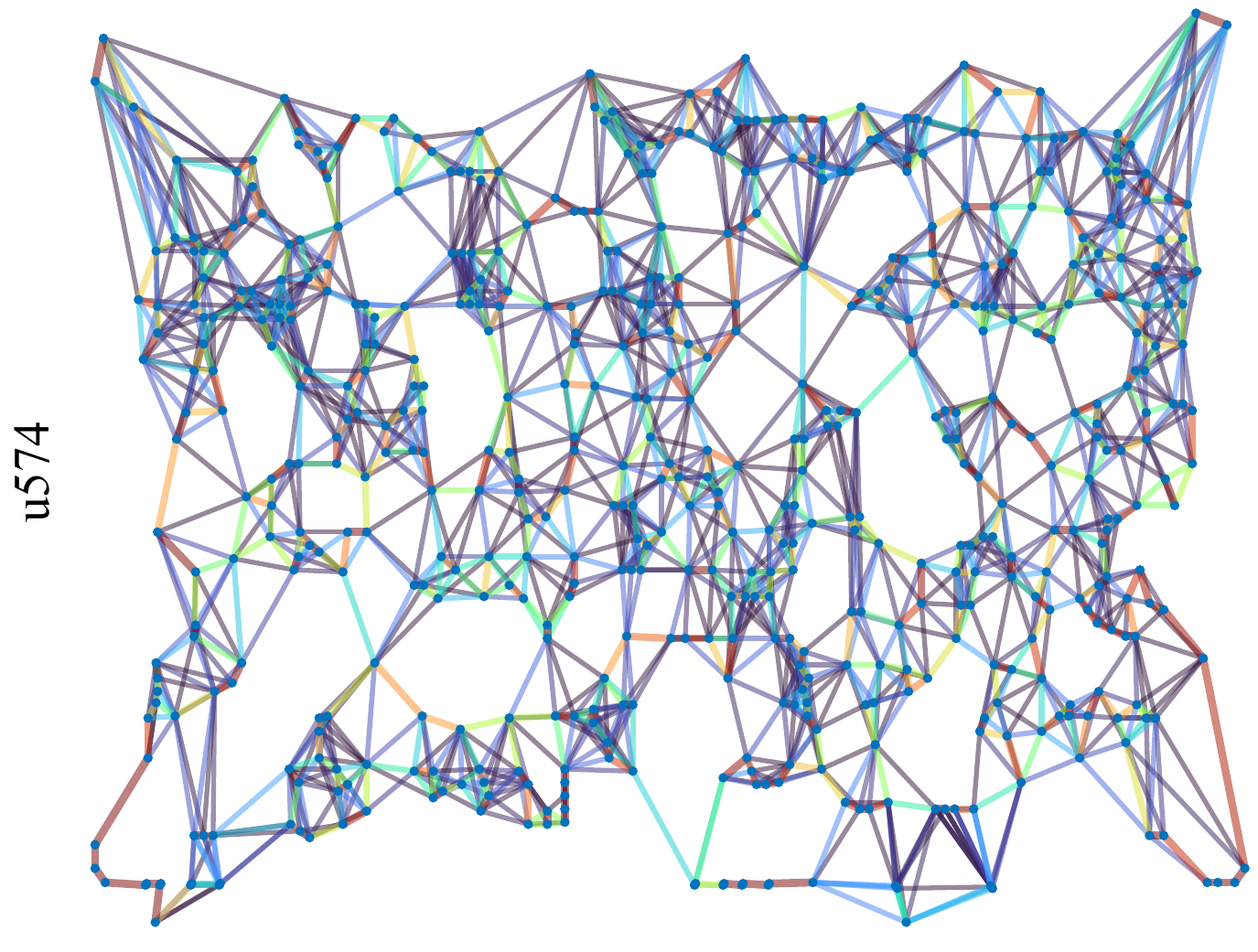}
\hskip5pt
\includegraphics[width=.23\columnwidth]{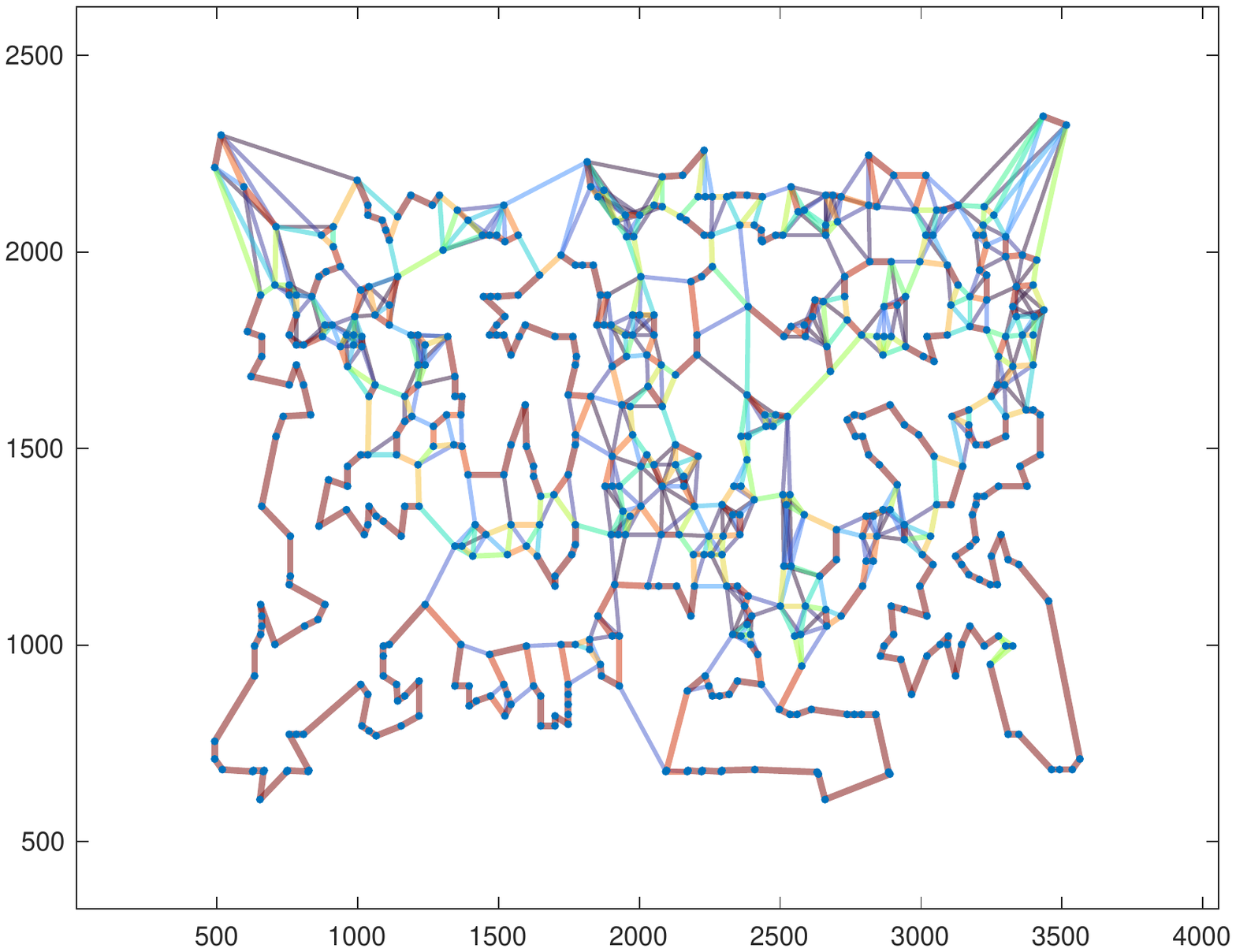}
\hskip5pt
\includegraphics[width=.23\columnwidth]{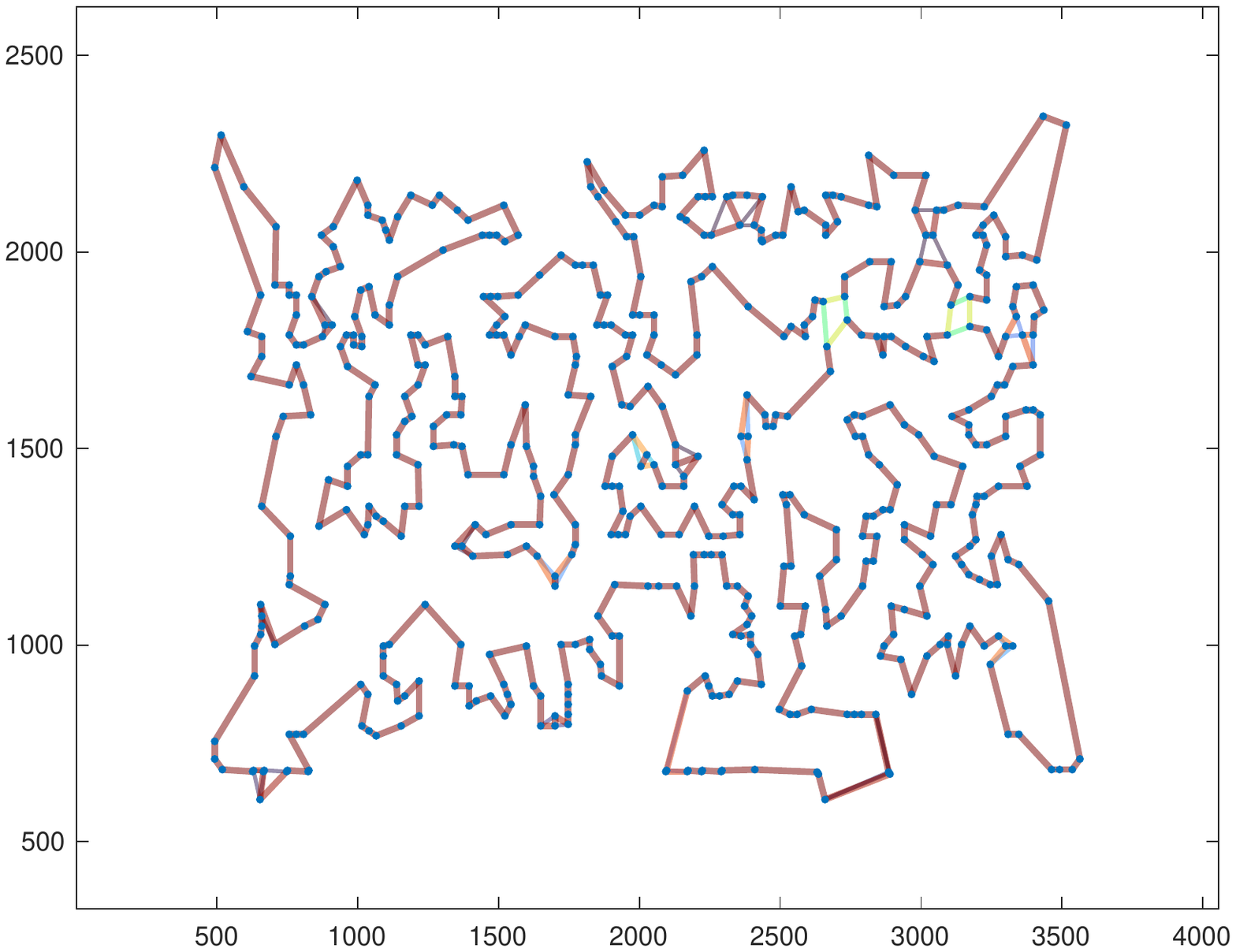}
\hskip5pt
\includegraphics[width=.23\columnwidth]{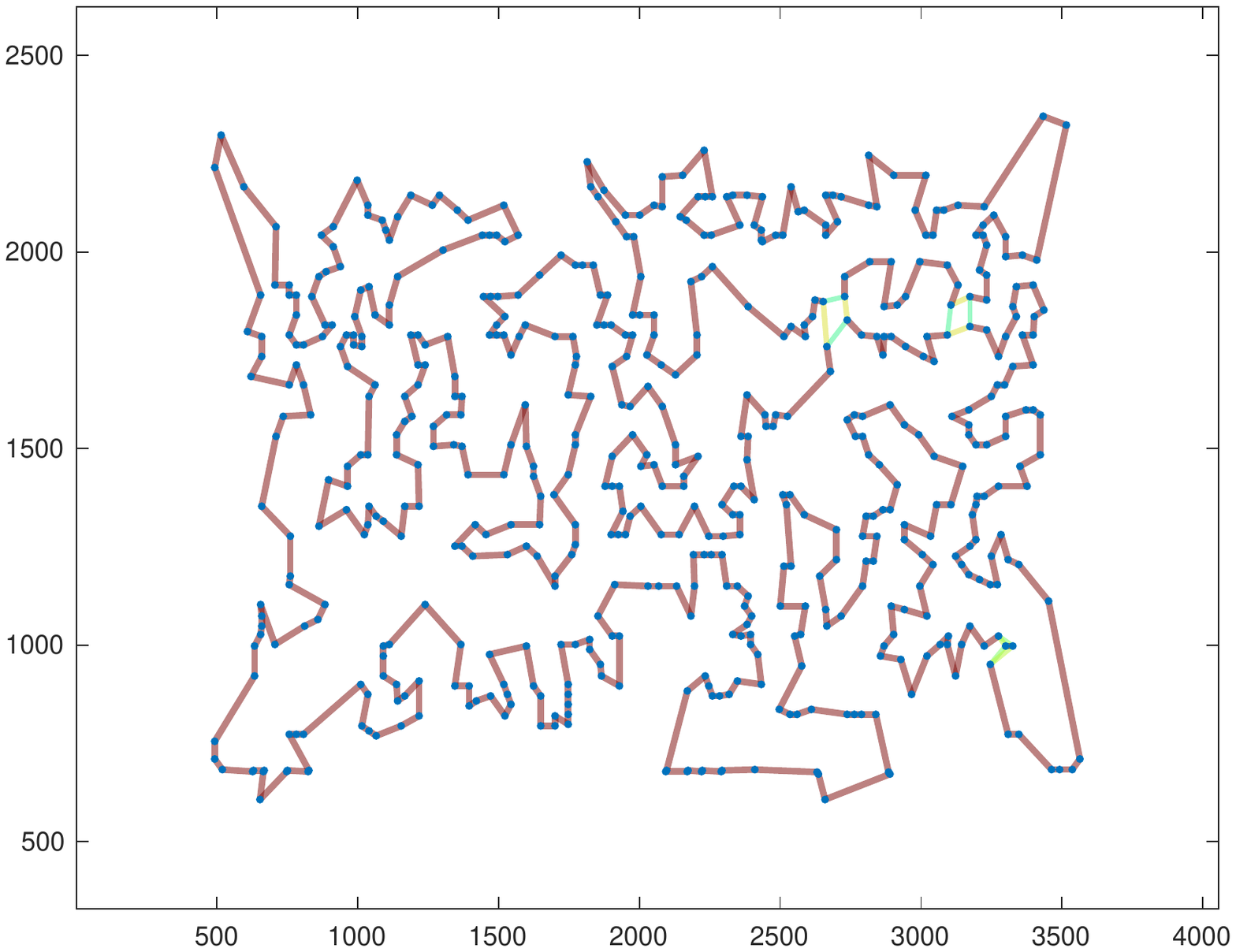}
\hskip15pt
\begin{tikzpicture}
\node[opacity=.55] (legend) at (0,0) {\includegraphics[width=0.6\columnwidth,trim=20pt 40pt 10pt 0,clip]{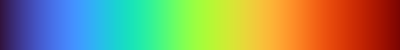}};
\node (low) at (-3.6,-0.3) {\scriptsize{\textcolor{gray!90}{low}}};
\node (medium) at (0,-0.3) {\scriptsize{\textcolor{gray!90}{medium}}};
\node (high) at (3.6,-0.3) {\scriptsize{\textcolor{gray!90}{high}}};
\end{tikzpicture}
\caption{Overlay of all edges used in exemplary final populations. Edges are colored by their frequency.}
\label{fig:edge_overlays}
\end{figure*}
Figure~\ref{fig:edge_overlays} visualises exemplary populations obtained by the single-stage and the two-stage EAX-EDO, Gurobi and EAX. The figure aids to comprehend how populations obtained by the EAX-EDO algorithms differ from the ones computed by standard EAX and Gurobi. As one can notice from Figure~\ref{fig:edge_overlays}, the single-stage EAX-EDO incorporates a higher number of edges into the population compared to the other algorithms. For example, on eil101, the population obtained from the single-stage EAX-EDO includes $758$ unique edges, while the number of edges for two-stage EAX-EDO, Gurobi and EAX are $416$, $238$, and $238$, respectively. A similar pattern can be observed on all the other instances; this includes a280 and u575, as it is shown in Figure~\ref{fig:edge_overlays}. Moreover, the figure depicts that EAX and Gurobi are almost incapable of having low frequent edges in the populations. However, the two EDO frameworks, especially those of the single-stage EAX-EDO, incorporate many low frequent edges into the populations. 

\subsubsection{Robustness of the populations}

One motivation for EDO is, as stated earlier, that decision-makers can choose between different alternatives if they are provided with a diverse set of high-quality solutions. For instance, decision-makers can avoid a certain edge if they prefer to, or the edge becomes unavailable for some reason. In this section, we compare the robustness of the population obtained from the four competitors when one or couples of edges of the optimal tour for a given TSP instance suddenly become unavailable. To this end, we randomly make one, two, and three edges of the optimal solution unavailable. Next, we determine 1) the percentage of occasions (over 1000 independent experiments) where there is at least one alternative tour in the population (encoded by $a$) 
and 2) the mean of different alternative tours in populations that avoid those edges (encoded by $d$). Table~\ref{tab:Robustness_test} summarises the results of this series of experiments.

\begin{table*}[t]
\caption{Comparison of the robustness of the populations obtained from single-stage EAX-EDO (1), two-stage EAX-EDO (2), EAX (3), and Gurobi (4) in case one, two, or three random edges from the optimal solution become unavailable in 100 runs. $a$ denotes the percentage of times the population has at least one alternative for the eliminated edges, while $d$ represents the number alternative tours avoiding the eliminated edges on average.}
\label{tab:Robustness_test}
\centering
\renewcommand{\tabcolsep}{1.3pt}
\renewcommand{\arraystretch}{1.6}
\begin{footnotesize}
\begin{tabular}{lcccccccccccccccccccccccc}
\toprule
& \multicolumn{8}{c}{\textbf{One edge}} & 
\multicolumn{8}{c}{\textbf{Two edges}} &
\multicolumn{8}{c}{\textbf{Three edges}} \\
\cmidrule(l{2pt}r{2pt}){2-9}
\cmidrule(l{2pt}r{2pt}){10-17}
\cmidrule(l{2pt}r{2pt}){18-25}
& \multicolumn{2}{c}{\textbf{EDO (1)}}
& \multicolumn{2}{c}{\textbf{EDO (2)}}
& \multicolumn{2}{c}{\textbf{EAX (3)}}
& \multicolumn{2}{c}{\textbf{Gurobi (4)}}
& \multicolumn{2}{c}{\textbf{EDO (1)}}
& \multicolumn{2}{c}{\textbf{EDO (2)}}
& \multicolumn{2}{c}{\textbf{EAX (3)}}
& \multicolumn{2}{c}{\textbf{Gurobi (4)}}
& \multicolumn{2}{c}{\textbf{EDO (1)}}
& \multicolumn{2}{c}{\textbf{EDO (2)}}
& \multicolumn{2}{c}{\textbf{EAX (3)}}
& \multicolumn{2}{c}{\textbf{Gurobi (4)}}\\
\cmidrule(l{2pt}r{2pt}){2-3}
\cmidrule(l{2pt}r{2pt}){4-5}
\cmidrule(l{2pt}r{2pt}){6-7}
\cmidrule(l{2pt}r{2pt}){8-9}
\cmidrule(l{2pt}r{2pt}){10-11}
\cmidrule(l{2pt}r{2pt}){12-13}
\cmidrule(l{2pt}r{2pt}){14-15}
\cmidrule(l{2pt}r{2pt}){16-17}
\cmidrule(l{2pt}r{2pt}){18-19}
\cmidrule(l{2pt}r{2pt}){20-21}
\cmidrule(l{2pt}r{2pt}){22-23}
\cmidrule(l{2pt}r{2pt}){24-25}
& $a$ & $d$ & $a$ & $d$& $a$ & $d$& $a$ & $d$& $a$ & $d$& $a$ & $d$& $a$ & $d$& $a$ & $d$& $a$ & $d$& $a$ & $d$& $a$ & $d$& $a$ & $d$\\
\midrule
eil101&\hl{\textbf{90}}&\hl{\textbf{18.07}}&50&9.5&18&4.34&18&3.79&\hl{\textbf{74}}&\hl{\textbf{6.57}}&20&1.82&3&0.33&3&0.27&\hl{\textbf{50}}&\hl{\textbf{2.29}}&8&0.43&1&0.04&0&0.03\\
a280&\hl{\textbf{83}}&\hl{\textbf{15.15}}&55&7.58&23&3.83&26&1.89&\hl{\textbf{64}}&\hl{\textbf{5.03}}&21&1.21&5&0.34&3&0.18&\hl{\textbf{40}}&\hl{\textbf{1.58}}&7&0.25&1&0.04&0&0.01\\
pr439&\hl{\textbf{82}}&\hl{\textbf{14.95}}&36&5.05&0&0&10&0.70&\hl{\textbf{58}}&\hl{\textbf{4.41}}&7&0.40&0&0&0&0.01&\hl{\textbf{30}}&\hl{\textbf{1.20}}&2&0.07&0&0&0&0\\
u574&\hl{\textbf{88}}&\hl{\textbf{15.37}}&46&6.79&1&0.27&6&0.42&\hl{\textbf{65}}&\hl{\textbf{5.08}}&15&1.13&0&0&0&0.01&\hl{\textbf{39}}&\hl{\textbf{1.58}}&5&0.252&0&0&0&0\\
rat575&\hl{\textbf{85}}&\hl{\textbf{14.40}}&61&9.06&26&4.64&21&0.94&\hl{\textbf{57}}&\hl{\textbf{4.08}}&29&1.92&4&0.391&1&0.02&\hl{\textbf{33}}&\hl{\textbf{1.30}}&9&0.36&1&0.04&0&0\\
p654&\hl{\textbf{90}}&\hl{\textbf{23.84}}&61&14.19&55&10.24&33&8.20&\hl{\textbf{79}}&\hl{\textbf{11.48}}&39&4.15&27&2.16&9&1.89&\hl{\textbf{70}}&\hl{\textbf{6.08}}&24&1.28&11&0.49&2&0.26\\
rat783&\hl{\textbf{82}}&\hl{\textbf{13.54}}&55&7.33&19&2.02&4&0.65&\hl{\textbf{56}}&\hl{\textbf{3.98}}&19&1.18&3&0.12&0&0.016&\hl{\textbf{32}}&\hl{\textbf{1.24}}&7&0.21&1&0.006&0&0\\
\bottomrule
\end{tabular}
\end{footnotesize}
\label{tab:Robustness_tes}
\end{table*}

The outcome indicates that the two variants of EAX-EDO lead to more robust populations against minor changes compared to EAX and Gurobi. In fact, the single-stage EAX-EDO has superior performance compared to its competitors: its population, e.g. on a280, succeeds in $83\%$ of occasions to offer an alternative when an edge becomes unavailable, whereas the Two-stage EAX-EDO scores $55\%$,  the EAX scores $23\%$ and the Gurobi achieves $26\%$, respectively. Moreover, the population of the single-stage EDO includes $5.8$ alternative tours to the optimal on average when two edges become unavailable. This is while, this figure is $1.69$ for the Two-stage EAX-EDO, and the populations of EAX and the Gurobi optimiser are incapable of offering any alternative on two and three instances, respectively. In case three edges are eliminated, the standard EAX and Gurobi are barely able to offer any alternatives, while success rates of the single-stage and the two-stage EAX-EDO are $42\%$ and $8.86\%$, respectively. Therefore, considering Table~\ref{tab:Robustness_test}, we can claim that the single-stage EAX-EDO framework outperforms the classic and the two-stage EDO frameworks in terms of population robustness.


\section{Conclusion}
\label{sec:conclusion}


In this paper, we introduced EAX-based evolutionary diversity optimisation approaches for the well-known traveling salesperson problem~(TSP) which are able to compute diverse sets of high quality TSP tours.
We designed an entropy-based diversity measure for the TSP and modified the powerful edge assembly crossover~(EAX) operator towards a variant called EAX-EDO crossover that allows to simultaneously minimise the tour length and maximise the population diversity. The resulting EAX-EDO algorithms allow to compute high quality diverse sets of TSP tours through a two-stage approach that alternates between optimised tour lengths and the diversity of the population or a single-stage method that optimises both criteria simultaneously.
Our experimental results show that (1) EAX-EDO crossover outperforms recent approaches from the literature based on $k$-OPT neighborhood search in a setting where the optimal tour is known and (2) the introduced algorithms show superior performance with respect to diversity while being competitive with respect to the objective function in comparison with the pure EAX and the Gurobi optimiser on a subset of classical TSP benchmark instances when an optimal solution in unknown. Moreover, our results indicates that EAX-EDO algorithms compute more robust populations compared to the classic optimisation frameworks. 

Future work will focus on the enhancement of EAX-EDO in terms of (1) generating several offspring in each iteration, and (2) a method to select a population from a pool of new offspring and old individuals. We expect these modifications to boost the performance of EAX-EDO in terms of both quality and diversity. Moreover, it is intriguing to investigate the application of EDO on real-world problems.    

\section*{acknowledgements}
This work was supported by the Australian Research Council~(ARC) through grants DP190103894 and FT200100536, and by the South Australian Government through the Research Consortium "Unlocking Complex Resources through Lean Processing".


\bibliographystyle{ACM-Reference-Format}
\bibliography{bib}

\end{document}